\documentclass[letterpaper, 10 pt, conference]{ieeeconf}
\IEEEoverridecommandlockouts % This command is only needed if
\makeatletter
\let\NAT@parse\undefined
\makeatother
\usepackage[numbers]{natbib}
\usepackage{placeins}

\usepackage{amsmath} % assumes amsmath package installed
\usepackage{amssymb} % assumes amsmath package installed
\usepackage{bm}
\usepackage{graphicx}
\usepackage{mathtools}
\usepackage{hyperref}
\usepackage{cleveref}
\usepackage{siunitx}
\usepackage{tikz}
\usetikzlibrary{shapes}
\usepackage{multirow}
\usepackage{makecell}
\usepackage{cuted}
\usepackage{textcomp}
\usepackage{stfloats}
\usepackage{url}
\usepackage{verbatim}
\usepackage{cite}
\usepackage{gensymb}

\usepackage{subcaption}
\usepackage{tablefootnote}
\usepackage{bm}
\usepackage{algorithm}
\usepackage[noend]{algpseudocode}
\usepackage{multirow}
\usepackage{subcaption}
\usepackage{cleveref} 
\usepackage{setspace}

\DeclareMathOperator*{\argmax}{arg\,max}

\newcommand{\reviewPrev}[1]{{\color{black}{#1}}}
\newcommand{\review}[1]{{\color{black}{#1}}}

\Crefformat{figure}{#2Fig.~#1#3}
\Crefmultiformat{figure}{Figs.~#2#1#3}{ and~#2#1#3}{, #2#1#3}{ and~#2#1#3}
\Crefformat{equation}{#2Eq.~#1#3}

\usepackage{xspace}
\makeatletter
\DeclareRobustCommand\onedot{\futurelet\@let@token\@onedot}
\def\@onedot{\ifx\@let@token.\else.\null\fi\xspace}
\makeatother

\DeclareRobustCommand{\revision}[1]{{\textcolor{black}{#1}}}

\definecolor{forestgreen}{RGB}{34,139,34}
\definecolor{orange}{RGB}{255, 191, 0}

\newcommand{\maxK}{N}
\newcommand{\maxJ}{M}

\newcommand{\maxF}{T}
\newcommand{\maxSeg}{S}

\newcommand{\overhorizon}[2]{#1^{#2}}
\newcommand{\indexK}[2]{#1_{#2}}

\newcommand{\systembf}[2]{{}^{\mathcal{#2}}{\mathbf{#1}}}
\newcommand{\system}[2]{{}^{\mathcal{#2}}{#1}}
\newcommand{\barbf}[1]{\mathbf{\bar{#1}}}
\newcommand{\alphabf}{{\bm{\alpha}}}

\newcommand{\ac}{\mathcal{A}}

\newcommand{\qb}{\mathbf{q}_\text{b}}
\newcommand{\qj}{\mathbf{q_j}}
\newcommand{\q}{\mathbf{q}}

\newcommand{\X}{\mathbf{x}^{0:\maxF_\alphabf}}

\newcommand{\Xtar}{{\mathbf{\bar{x}}}^{0:T}}

\newcommand{\Qtar}{\bar{\q}^{0:T}}
\newcommand{\x}{\mathbf{x}}
\newcommand{\U}{{\mathbf{u}}^{0:\maxF_\alphabf}}
\newcommand{\Keymotion}{\p^{0:\maxF}}

\newcommand{\p}{\mathbf{p}}
\newcommand{\jp}{\bm{\theta}}
\newcommand{\PI}[1]{\mathcal{P}(#1)}
\newcommand{\e}{\mathbf{e}}
\newcommand{\pb}{\mathbf{p}_\text{b}}
\newcommand{\h}{\mathbf{h}}
\newcommand{\con}{\mathbf{c}}
\newcommand{\FK}{\text{FK}}
\newcommand{\LI}{\text{LI}}
\newcommand{\qsmr}{\bar{\q}}

\newcommand{\quv}{\system{\q}{\text{IK}}}
\newcommand{\dquv}{\system{\dot{\q}}{\text{IK}}}

\newcommand{\vexit}{\mathbf{v}_\text{exit}}

\newcommand{\qref}{\tilde{\q}}
\newcommand{\dqref}{\delta\tilde{\q}}
\newcommand{\kernel}{\mathbf{K}}

\newcommand{\optimal}[1]{{}^{*}{#1}}

\newcommand{\MM}{\mathrm{mm}}

\definecolor{lightblue}{HTML}{9BB7D4} % cerulean
\definecolor{light2blue}{HTML}{92A8D1} % serenity
\definecolor{aqua}{HTML}{7BC4C4} % aqua sky
\definecolor{turquoise}{HTML}{53B0AE} % blue turquoise
\definecolor{iris}{HTML}{5A5B9F} % blue iris

\definecolor{darkyellow}{HTML}{F0C05A} % mimosa
\definecolor{yellow}{HTML}{F8D948} % illuminating

\definecolor{pink}{HTML}{D94F70} % honeysuckle
\definecolor{lightpink}{HTML}{F7CAC9} % rose quartz
\definecolor{warmpink}{HTML}{FF6F61} % living coral
\definecolor{darkpink}{HTML}{C74375} % fuchsia rose
\definecolor{magenta}{HTML}{BB2649} % viva magenta

\definecolor{orange}{HTML}{E2583E} % tigerlily
\definecolor{lightorange}{HTML}{FEBE98} % peach fuzz
\definecolor{orangered}{HTML}{DD4124} % tangerine tango

\definecolor{green}{HTML}{009473} % emerald
\definecolor{grassgreen}{HTML}{88B04B} % greenery

\definecolor{redbrown}{HTML}{955251} % marsala
\definecolor{darkred}{HTML}{9B1B30} % chili pepperV

\definecolor{purple}{HTML}{6968AC} % violet, very peri
\definecolor{darkpurple}{HTML}{5F4B8B} % ultra violet
\definecolor{lightpurple}{HTML}{B163A3} % radiant orchid

\definecolor{grey}{HTML}{959A9C} % ultimate gray
\definecolor{beige}{HTML}{DECDBE} % sand dollar

\hypersetup{
    colorlinks=true, % false: boxed links; true: colored links
    linkcolor=black, % color of internal links
    citecolor=darkred, % color of links to bibliography
    filecolor=blue, % color of file links
    urlcolor=grey,
}

\title{\LARGE \bf \revision{Spatio-Temporal Motion Retargeting for Quadruped Robots} }

\author{Taerim~Yoon$^{1}$,
        Dongho~Kang$^{2}$,
        Seungmin~Kim$^{1}$,
        \review{Jin~Cheng$^{2}$},
        \review{Minsung~Ahn$^{3}$}, \\
        Stelian~Coros$^{2}$, and 
        Sungjoon~Choi$^{1}$
\thanks{Manuscript received September 17, 2024; 
\review{This work was supported by Institute of Information \& communications Technology Planning \& Evaluation (IITP) grant funded by the Korea government(MSIT) (No. RS-2019-II190079, Artificial Intelligence Graduate School Program Korea University, 12.5\%), (No. RS-2024-00457882, AI Research Hub Project, 12.5\%), (No. RS-2022-II220871, AI Autonomy and Knowledge Enhancement for AI Agent Collaboration, 12.5\%), (No. RS-2022-II220480, Development of Training and Inference Methods for Goal-Oriented Artificial Intelligence Agent, 12.5\%), (No. RS-2024-00336738, Development of Complex Task Planning Technologies for Autonomous Agents, 50\%).
Additionally, this work utilized research resources sponsored by the European Research Council (ERC) under the European Union’s Horizon 2020 research and innovation programme (grant agreement No. 866480).
}
(\emph{Corresponding authors: Sungjoon Choi.)}
}
\thanks{$^{1}$Taerim~Yoon, Seungmin~Kim, and Sungjoon~Choi are with the Department of Artificial Intelligence, Korea University, 145 Anam-ro, Seongbuk-gu, Seoul, Korea (email: taerimyoon@korea.ac.kr; dkslanjdi96@korea.ac.kr; sungjoon-choi@korea.ac.kr).}
\thanks{$^{2}$Dongho~Kang, Jin~Cheng and Stelian Coros are with the Department of Computer Science, ETH Zurich, Wasserwerkstrasse 12, 8092 Zurich, Switzerland (email: kangd@ethz.ch, jicheng@ethz.ch, scoros@ethz.ch).}
\thanks{$^{3}$Minsung~Ahn is with the Department of Mechanical and Aerospace Engineering, UCLA, 420 Westwood Plaza, Los Angeles, 90095, CA, USA (email: aminsung@ucla.edu)}
}

\begin{document}
    \maketitle
    \pagestyle{plain}

    %%%%%%%%%%%%%%%%%%%%%%%%%%%%%%%%%%%%%%%%%%%%%%%%%%%%%%%%%%%%%%%%%%%%%%%%%%%%%%%%
    \begin{abstract}
    This work presents a motion retargeting approach for legged robots, \reviewPrev{aimed at transferring the dynamic and agile movements to robots from source motions}. 
    In particular, we guide the imitation learning procedures by transferring motions from source to target, effectively bridging the morphological disparities while ensuring the physical feasibility of the target system.
    In the first stage, we focus on motion retargeting at the kinematic level by generating kinematically feasible whole-body motions from keypoint trajectories.
    Following this, we refine the motion at the dynamic level by adjusting it in the temporal domain while adhering to physical constraints.
    This process facilitates policy training via reinforcement learning, enabling precise and robust motion tracking. 
    We demonstrate that our approach successfully transforms \reviewPrev{noisy motion sources}, such as hand-held camera videos, into robot-specific motions that align with the morphology and physical properties of the target robots.
    \reviewPrev{Moreover, we demonstrate terrain-aware motion retargeting to perform BackFlip on top of a box.}
    We successfully deployed these skills to four robots with different dimensions and physical properties in the real world through hardware experiments.
    \end{abstract}

    %%%%%%%%%%%%%%%%%%%%%%%%%%%%%%%%%%%%%%%%%%%%%%%%%%%%%%%%%%%%%%%%%%%%%%%%%%%%%%%%
    % sections
    \section{Introduction}
Legged robots are steadily making their way into human society for their ability to walk alongside humans. 
As these robots become more prevalent in everyday settings, there is growing interest in generating natural and subtle motions beyond standard walking~\citep{kang_animal_2022, cat_like_jumping_tro}.
In this context, \emph{imitation learning} (IL) has emerged as an effective tool for generating natural motion by imitating prerecorded or hand-crafted motions~\citep{serifi2024vmp}.
For example, safe and attentive behaviors of robotic assistance dogs can be developed by imitating the motions of a service dog captured in a video, where the dog carefully approaches the elderly without causing disruption.

\begin{figure}[!t]
    \centering
    \includegraphics[width=0.99\linewidth]{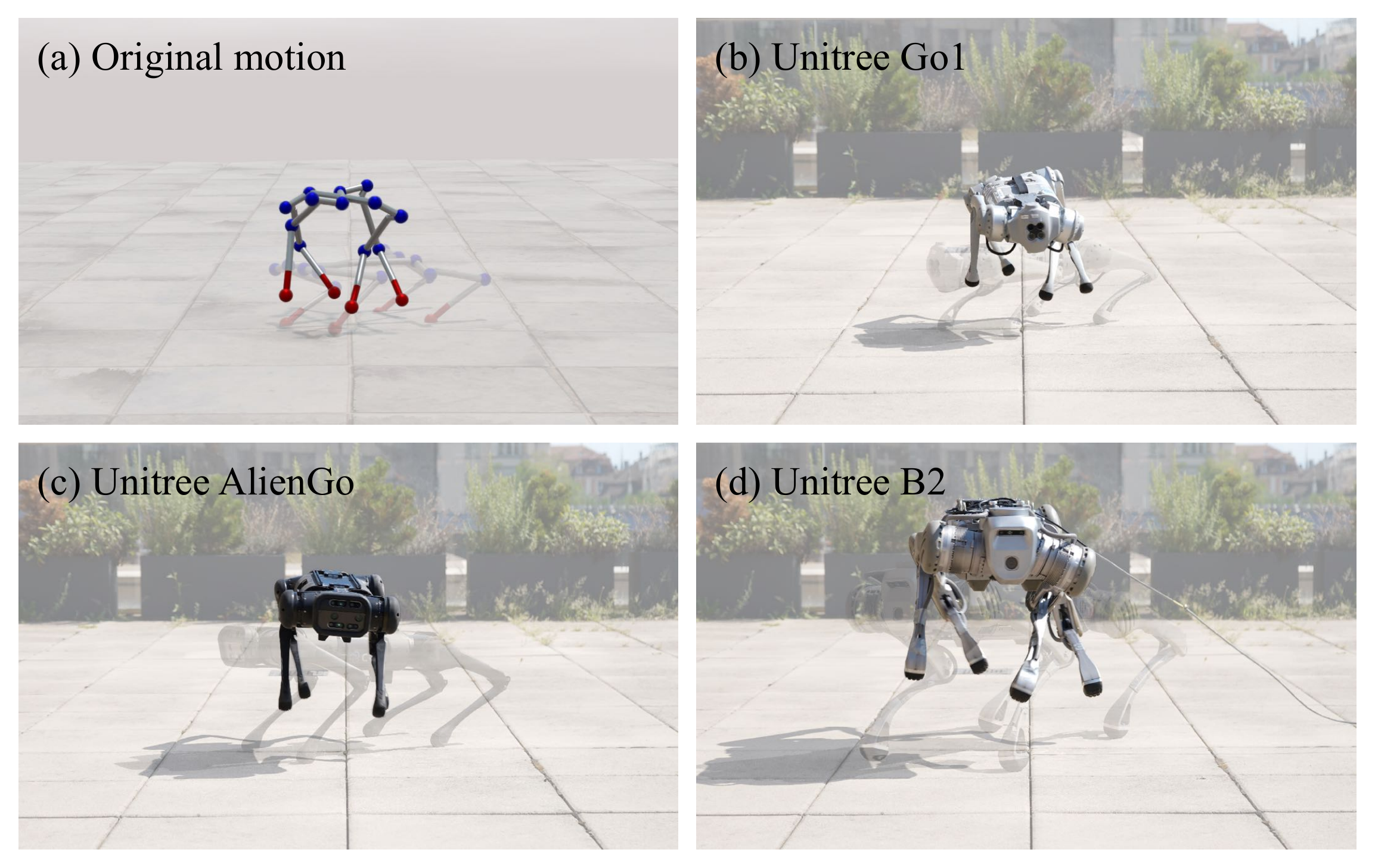}
    \caption{(a) A hand-crafted HopTurn motion was kino-dynamically retargeted using our method and executed on the (b) \emph{Unitree Go1}, (c) \emph{AlienGo}, and (d) \emph{B2} robots, each with different dimensions and physical properties.}
    \label{fig:teaser}
\end{figure}

% Challenges-1
% topic sentence: must overcome morphology difference. 
The main challenge of motion imitation lies in overcoming the morphological and dimensional differences between source and target systems~\citep{feng2023genloco, simulataneous_motion_optimize}. 
Specifically, when imitating prerecorded animal motion, the disparity between the animal actor and the robot in morphological and physical properties hampers the direct transfer of the motion at the joint trajectory levels.
To address this issue, motion imitation involves a process known as \emph{motion retargeting}, which adjusts the target motion to ensure compatibility with the size and morphology of the target robotic system.

% Challenges-2
% topic sentence: intro to motion retargeting and it's current limitation
Existing motion retargeting methods can transfer and adapt source motions for target systems but often produce kino-dynamically infeasible motions. 
\reviewPrev{These infeasible motions can result in sub-optimal mimicking behaviors or even complete failure in imitation~\citep{ mpc_data_aug_tro, data_template_tro}}. 
Additionally, the application of these methods is generally limited to motion data that includes whole-body movements with global body pose information. Consequently, this limitation restricts the use of motion data with an unknown coordinate frame, such as animal movements captured by a hand-held camera.

% Goal
% topic sentence: our goal is imitate source motions for natural behaviors 
Our work aims to generate physically feasible reference motions to facilitate streamlined and successful learning of control policies that imitate the expressiveness and agility in source movements, as shown in \Cref{fig:teaser}. 
More specifically, we aim to develop a motion retargeting method that enables transferring motion data lacking global body pose information and with an unknown origin point to target robotic systems. 

\begin{figure*}[!t]
    \centering
    \captionsetup[subfloat]{labelfont=normal, font=normal}
    \subfloat[{Overview}]{
        \includegraphics[width=0.95\linewidth]{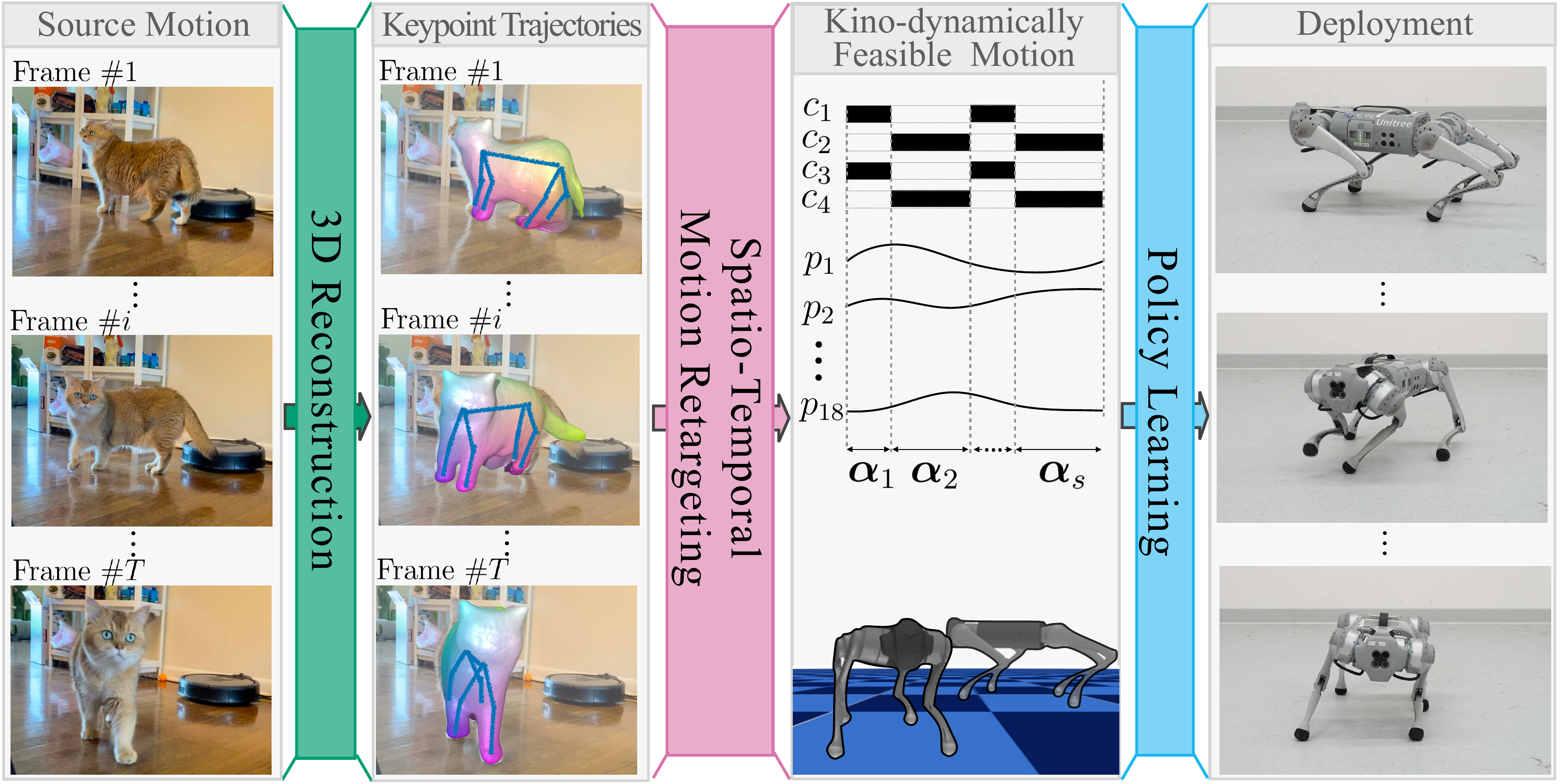}
        \label{fig:overview}
    } \\
    \vspace{0.1cm}
    \subfloat[{Spatial motion retargeting}]{
        \includegraphics[width=0.55\linewidth]{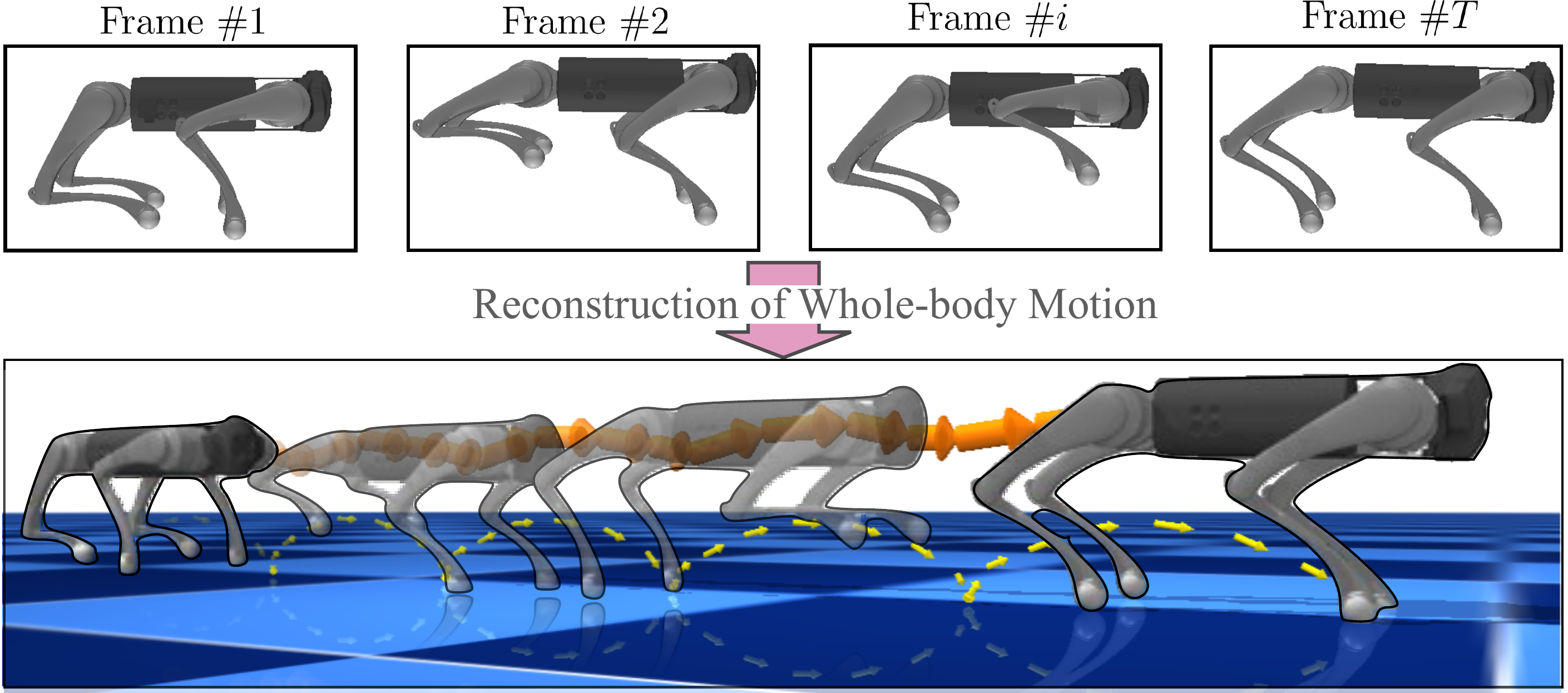}
        \label{fig:SMR}
    } 
    \subfloat[{Temporal motion retargeting}]{
        \includegraphics[width=0.35\linewidth]{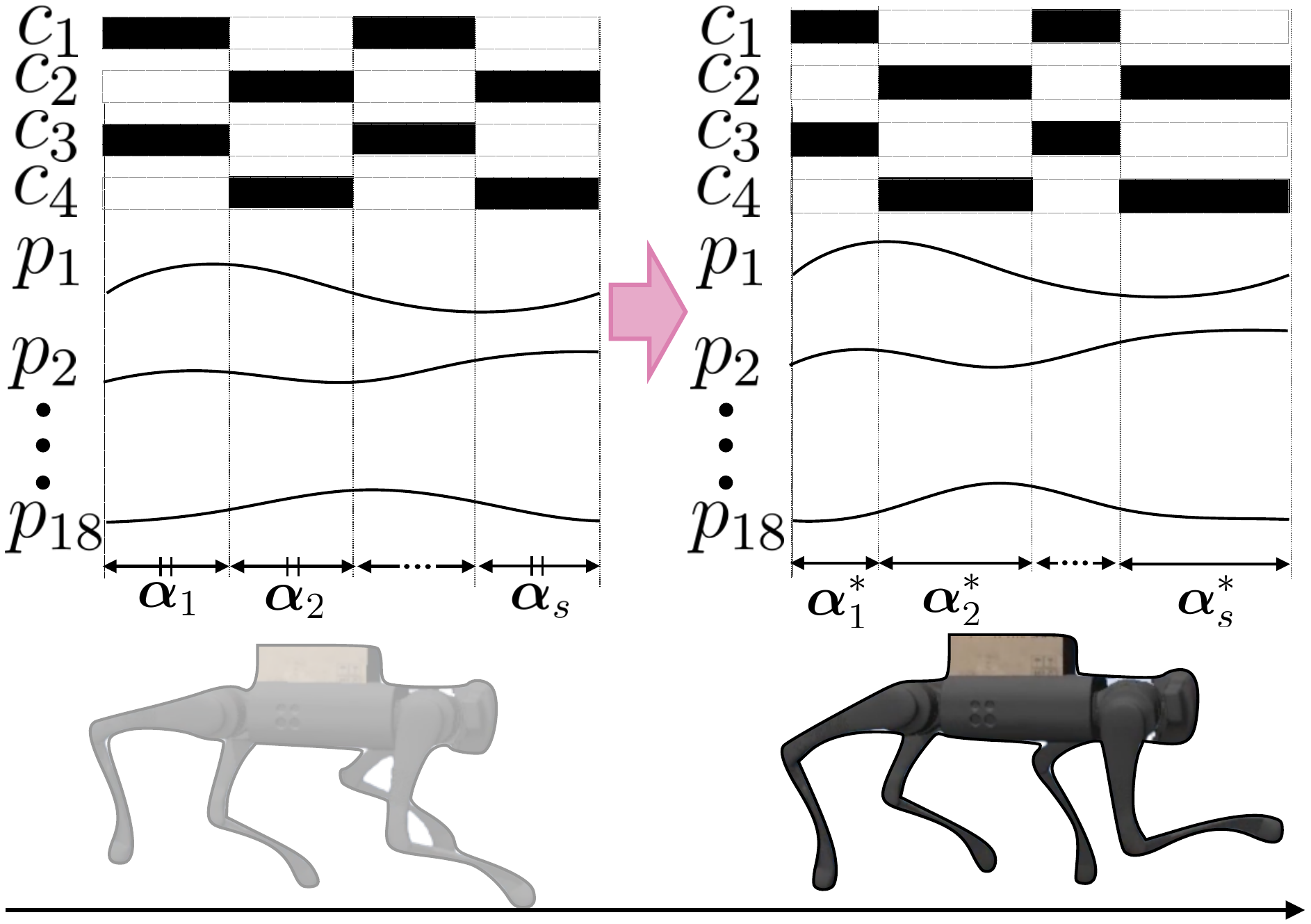}
        \label{fig:TMR}
    } 
    \caption{(a) 
    Our STMR method consists of SMR and TMR stages to generate kino-dynamically feasible motion. (b) In the SMR stage, a kinematically feasible whole-body motion in absolute coordinates is generated, but only keypoint motion is given in local coordinates. (c) In the TMR stage, the temporal aspect of the motion is optimized, and a dynamically feasible motion is generated. As the resulting motion is physically feasible, it can guide the training of control policy toward successful deployment in the real world.}
    \label{fig:overview_total}
\end{figure*}

% topic sentence: presentation of our approach.
To this end, we propose \emph{spatio-temporal motion retargeting} (STMR), which 
transfers baseless keypoint trajectories to the target robot as shown in \Cref{fig:overview}.
\reviewPrev{The motivation of our approach is to break down motion retargeting problems into two subproblems in space and time domains, respectively.}
In more detail, STMR generates whole-body motion with two sequential processes, namely \emph{spatial motion retargeting} (SMR) and \emph{temporal motion retargeting} (TMR).
\reviewPrev{SMR retargets motion at a kinematic level.
By regulating kinematic artifacts of foot sliding and foot penetrations, SMR enables the generation of whole-body motion from videos by reconstructing it from baseless keypoint trajectories, as depicted in \Cref{fig:SMR}.
}
On the other hand, TMR, as illustrated in \Cref{fig:TMR}, focuses on refining the motion subject to dynamics constraints and further deforms the motion in the temporal domain to generate a dynamically feasible motion.
This step is particularly crucial for motions that involve flight phases, as motions in \Cref{fig:teaser}, where variations in robot size and actuation power should lead to differences in mid-air duration.
In the final step, a feedback control policy is trained through reinforcement learning (RL), guided by a kino-dynamically feasible reference motion to ensure accurate and robust tracking when deployed on real robots.

% topic sentence: summary of results
To demonstrate the efficacy of our approach, we conducted extensive experiments with distinct motions across various quadrupedal robot platforms and compared the results with three baseline methods for motion imitation.
Additionally, we quantitatively show that motions generated by our STMR method are free of foot sliding and preserve contact schedules.
We also showcase that STMR can generate whole-body motion from the relative movement of keypoints and contact schedules, which we refer to as baseless motion.
Finally, we demonstrate that a learned control policy can be successfully deployed in the real world on four robots---\emph{Unitree Go1}, \emph{Unitree Go2}, \emph{AlienGo}, and \emph{B2}---each with different dynamic properties and dimensions.

% topic sentence: summary of the main contribution 
In summary, we propose STMR, which generates a kino-dynamically feasible motion from keypoint trajectories described relative to an unknown reference coordinate frame and facilitates successful IL.
The key contributions of our work are summarized as follows: 
\begin{enumerate}
    \item We introduce spatial and temporal motion retargeting (STMR) that transfers motion by adjusting in both spatial and temporal dimensions to ensure the physical feasibility of motion imitation.
    \item We present a novel nested optimization framework for temporal motion retargeting, which integrates a model-based controller as an internal process to optimize motion timing.
    \item We experimentally show that the motion optimized by STMR leads to successful policy learning for real-world deployment.
\end{enumerate}

%%%%%%% Related Work %%%%%%%%%%
\section{Related Work} 
\label{Sec:RelatedWork}
\subsection{Motion imitation for quadruped robots}
% topic sentence: Historical context of motion imitation
Developing a legged locomotion controller capable of replicating the agile and natural movements of legged animals has been a longstanding aspiration. 
To realize this ambition, a body of research has explored the approach of incorporating prerecorded animal motion or hand-crafted motion animation into a legged locomotion control pipeline.

% topic sentence: Motion imitation with MBOC
Several studies have demonstrated motion imitation with a model-based legged locomotion control pipeline. 
\citet{kang_animal_2022} employed a simplified dynamics model and a gradients-based optimization method to search for the control sequence, including the footholds of robots, in order to transfer the gait sequences that maintain the non-periodic and irregular patterns of animal motion. 
\citet{grandia_doc_2023} introduced a nested optimization approach for the retargeting of animal motion by deriving sensitivities in the retargeting parameters, with the goal of creating dynamically feasible target motions. These motions were then executed on a quadruped robot using model predictive control. 
\citet{slowmo} presented a motion imitation pipeline that transfers source motions obtained from videos using a pose estimator. Additionally, they adopted the contact implicit trajectory optimization technique to remove the requirement for explicit contact information in the motions.
Notably, \citet{kang_animal_2021} demonstrated a model-based motion controller that follows high-level joystick commands while preserving animal-like walking styles by incorporating a data-driven motion planning algorithm into the legged locomotion control pipeline.

% topic sentence: Motion imitation with imitation learning
In another vein, imitation learning (IL) has also been a vigorously explored area of research and represents a promising strategy for imitating animal motions. 
\citet{peng_deepmimic_2018} introduced a reinforcement learning (RL) approach that uses a reward function to align the state of a character with prerecorded motion data, enabling the character to perform actions such as walking, running, and dancing. 
This methodology was further extended to real-world robotics in later work, showcasing the execution of agile animal movements on a quadruped robot~\citep{peng2020learning}. 
More recently, the adversarial motion prior (AMP)~\citep{peng_amp_2021} method was introduced for improved generality and applied for a quadruped robot to walk in a real-world scenario~\citep{escontrela_adversarial_2022}. 
Inspired by this, \citet{li2023learning} adopted a similar approach to imitate rough and physically infeasible reference torso motions.

% topic sentence: MBOC + IL
% rather than relying on motion data from real-world demonstrations,
Shifting the focus to animal motion imitation, some studies combine model-based optimal control (MBOC) and IL by leveraging MBOC demonstrations to train RL policies, resulting in dynamic and agile quadruped motions~\citep{levine2013guided, fuchioka_opt-mimic_2023, kang_rl_2023, mpc_data_aug_tro, data_template_tro}. Notably, \citet{fuchioka_opt-mimic_2023} employed offline trajectory optimization (TO) to generate complex reference motions such as quadruped backflipping and executed these motions using a feedback controller trained with IL. Similarly, \citet{kang_rl_2023} introduces on-demand reference motion generation through optimal control for efficient and robust IL across various quadruped gait patterns. \reviewPrev{\citet{liu2024opt2skill} utilized DDP to refine motion skills led by policy training.}

% topic sentence: We use MBOC and IL
In this work, we utilize prerecorded animal motion data and hand-crafted animations to replicate the agile and natural movements observed in animals. 
\reviewPrev{
Similar to the work by \citet{fuchioka_opt-mimic_2023}, \citet{kang_rl_2023} and \citet{liu2024opt2skill}, our approach enhances motion imitation by using MBOC to generate optimal control and state data for streamlined IL. 
However, the difference is that we not only optimize the control and states of the robot but also adjust the target motion, including temporal deformation.}

Additionally, we demonstrate that our method can retarget motions obtained from videos. 
This is similar to the work of \citet{slowmo} in that we refine motions obtained from pose-estimators according to the robot's dynamics. 
\reviewPrev{However, our proposed method differs in that it reconstructs the base trajectory of the robot in the global frame by incorporating contact plans, instead of relying on noisy base positions given by the pose-estimator.}
\vspace{-0.5em}

\subsection{Motion retargeting} % this section focus how previous work formulate "motion retargeting" and compare this to our method. 
% topic sentence: why need MR
In the context of motion imitation, overcoming the morphological differences between the source and target systems is essential to replicate motion from a system with distinct configurations.
In this regard, motion retargeting plays a crucial role by adapting the source motion to be compatible with the target system.

% topic sentence: MR - Paired data
The most intuitive motion retargeting methods often involve utilizing supervised learning with paired motion data between two source and target configurations. 
In line with this methodology, \citet{yamane} employed a Gaussian process latent model to map human motions to a character directly. 
\citet{seol_creature_2013} presented a technique of blending the retargeted motion with the nearest motion data point to efficiently learn the mapping for motion retargeting.
More recently, a mixture of experts were trained on a paired dataset to generate real-time quadruped motions~\citep{kim_humanconquad_2022}. 
Meanwhile, \citet{choi2020nonparametric} proposed a semi-supervised learning approach that constructs a latent space of collision-free poses and uses non-parametric regression to enable real-time motion retargeting in the real world. While these methods may appear straightforward, it is important to note that they often require a laborious data collection process, posing challenges in scaling the data for numerous configurations.

% topic sentence: Kinematic MR
Alternatively, another line of work focuses on retargeting motions at the kinematic level by transferring the movement of keypoint trajectories. 
The work by \citet{kwang-jin_choi_-line_1999} is among the first to utilize inverse kinematics (IK) for motion retargeting by following the keypoint trajectories.
\citet{sjchoi_iros_natual} further advanced this approach by modifying keypoint trajectories and employing IK to evaluate the deformed motion.

Building upon this foundation, the transfer of keypoint trajectories was further explored using unsupervised learning approaches.
\citet{villegas_neural_2018} leveraged the differentiability of forward kinematics to transfer motions between human-like characters by matching keypoint movements with adversarial loss.
Similarly, the work by \citet{10.1145/3610548.3618255} utilizes keypoint-wise feature loss and adversarial loss to retarget humanoid motions to non-humanoid characters. 
\citet{aberman_skeleton-aware_2020} introduced a concept of the common skeleton to construct an intermediate latent space shared among different kinematic structures using unsupervised learning. 
\citet{choi_self-supervised_2021} proposed a self-supervised learning framework to ensure a safe motion retargeting process, wherein pseudo-labels were acquired through optimization-based motion retargeting approaches.

% topic sentence: Dynamic motion retargeting
While the kinematic motion retargeting methods mentioned previously can generate visually convincing motions, incorporating system dynamics can significantly enhance the physical feasibility of motions and streamline their deployment to real robots.
\citet{tak_physically-based_2005} introduced the dynamic motion retargeting filter to regularize motion with Zero-Moment Point (ZMP) constraints for legged figures.
Following this, \citet{al_borno_robust_2018} employed linear quadratic regulator (LQR) search trees, and \citet{rouxel_multicontact_2022} used a whole-body optimization under kino-dynamic constraints to track keypoint trajectories of the source motion.
As previously mentioned, \citet{grandia_doc_2023} employed a nested optimization approach with MBOC to ensure the dynamical feasibility of the retargeted motions.

% topic sentence: Focus on Temporal optimization
In this paper, we transfer trajectories of local positions of keypoints from a source motion while preserving the contact schedule of the motion. Throughout this process, we account for both the kinematics and dynamics of the target system to generate physically feasible motions. Specifically, our method prevents kinematic artifacts (e.g., foot sliding) while adhering to the system's dynamics and physical constraints. Notably, we refine the motion in the temporal domain by adjusting the time scale. In our experiments, we demonstrate that this temporal optimization generates dynamically feasible motions, successfully transferring motions to a target system with significantly different dimensions.

\section{Preliminaries}
This section describes two established methods, namely the Unit Vector Method (UVM)~\citep{sjchoi_iros_natual} and Differential Dynamic Programming (DDP)~\citep{mayne_differential_1973}. 
Since our proposed method produces physically feasible movements by refining a target motion at kinematic and dynamic levels, these techniques are utilized as subprocesses: UVM for kinematic-level motion retargeting and DDP for dynamic-level motion retargeting.

\subsection{Unit vector method} 
\label{sec:unit vector}
The UVM retargets a source motion by preserving the directional unit vector between two adjacent keypoints that move along with the target robot.
While this method does not guarantee the kinematic feasibility of the retargeted motion, we use it as a subprocess to generate an initial reference for whole-body motion.

Consider a robot whose joint position is denoted as $\bm{\theta} \in \mathbb{R}^{\maxJ}$ and keypoint position as $\p \in \mathbb{R}^{\maxK \times 3}$, where $\maxJ$ and $\maxK$ are the numbers of joints and keypoints, respectively.
The Unit Vector Method aims to obtain joint position $\jp$ given the keypoint positions of the source system, denoted as $\systembf{\p}{\text{src}}$.

Let us define a parent index $\PI{j}$ for $j$th keypoint with respect to the kinematic tree.
The unit directional vector between $j$th keypoint and its parent can be described as $\e_j = (\p_j - \p_{\PI{j}})/d_j$ where $d_j := \|\p_j - \p_{\PI{j}}\|$ is constant as two keypoints lies on rigid link. 
By scaling the directional vector $\e_j$ with target link length $\system{d_j}{\text{trg}}$, the keypoint position of the target system $\systembf{\p}{\text{trg}}$ is obtained as
\begin{equation} \label{eq:unit vector method}
    \systembf{p}{\text{trg}}_j = \systembf{p}{\text{trg}}_{\PI{j}} + \system{d_j}{\text{trg}} \systembf{\e_j}{\text{src}}.
\end{equation}
Subsequently, the joint position $\jp$ is obtained by solving inverse kinematics:
\begin{equation*}
    \jp = \text{IK}(\systembf{p}{\text{trg}}_{1:\maxK}).
\end{equation*}

\subsection{Differential Dynamic Programming} \label{sec:ddp}
Differential Dynamic Programming (DDP) is an effective approach for finding control inputs that achieve user-defined objectives while satisfying the system dynamics model and physical constraints. 
As the proposed retargeting problem involves finding dynamically feasible motions, we utilize DDP as an internal subprocess for dynamic-level motion retargeting.
Specifically, we utilize an Iterative Linear Quadratic Gaussian (ILQG)~\citep{tassa_synthesis_2012}, a variant of DDP.

We denote the state of the robot as $\x$, the control input as $\mathbf{u}$, and the dynamics of the robot as $f$. The discrete-time dynamics at $i$th step is described as $\mathbf{x}^{i+1} = f(\mathbf{x}^{i},\mathbf{u}^{i}).$%Equation~\ref{eq:dynamics}.
\reviewPrev{
The primary goal of DDP is to find the optimal control inputs, $\mathbf{u}^{0:h-1}$, and states, $\mathbf{x}^{0:h}$, given the target states, $\Xtar$, under the system dynamics, $f$.
}
This can be represented as \Cref{eq:DDP_obj} where we minimize the objective function comprising the sum of the running cost $l_i$ and the final cost $l_f$:
\begin{equation} \label{eq:DDP_obj}
    \begin{aligned}
        \min_{\mathbf{x}^{0:h},\,\mathbf{u}^{0:h-1}} \quad & \sum_{i=0}^{h-1} l_i({}\mathbf{x}^{i},{}\mathbf{u}^{i}; {}\mathbf{\bar{x}}^{i}) + l_f(\mathbf{x}^{h}; \mathbf{\bar{x}}^{h})\\
        \text{s.t.} \quad & \overhorizon{\x}{i+1} = f(\overhorizon{\x}{i}, \mathbf{u}^i). \\
    \end{aligned}
\end{equation}

We define the optimal value function (also known as the optimal cost-to-go) at $i$th step as follows:
\begin{equation*}
    \begin{aligned}
        \optimal{V}^i(\mathbf{x}) = \min_{\mathbf{x}^{i:h},\,\mathbf{u}^{i:h-1}} \quad & \sum_{j=i}^{h-1} l({}\mathbf{x}^{j},\optimal{\mathbf{u}}^{j}; {}\mathbf{\bar{x}}^{j}) + l_f(\mathbf{x}^{h}; \mathbf{\bar{x}}^{h})\\
        \text{s.t.} \quad & \overhorizon{\x}{j+1} = f(\overhorizon{\x}{j}, \mathbf{u}^j)\\
        & \mathbf{x}^i = \mathbf{x}.
    \end{aligned}
\end{equation*}

According to the Bellman optimality principle, the relationship between the optimal value function at $i$ and $i+1$ steps can be expressed as
\begin{equation} \label{eq:value}
    \optimal{V}^{i}(\mathbf{x}) = \min_\mathbf{u}~[l_i(\mathbf{x}, \mathbf{u}) + \optimal{V}^{i+1}\big (f(\mathbf{x},\mathbf{u})\big)].
\end{equation}
Furthermore, we define the state-action value function $Q^i$, which is derived by perturbing the state-action pair $(\x^i, \mathbf{u}^i)$ around the minimum: 
\begin{equation*}
\label{eq:Q}
\begin{aligned}
Q^i(\mathbf{\delta x}, \mathbf{\delta u}) &= l_i(\mathbf{x} + \mathbf{\delta x}, \mathbf{u} + \mathbf{\delta u}) - l_i(\mathbf{x}, \mathbf{u}) \\
& \quad +  V^{i+1}(f(\mathbf{x} + \mathbf{\delta x}, \mathbf{u} + \mathbf{\delta u})) - V^{i+1}(f(\mathbf{x}, \mathbf{u})).
\end{aligned}     
\end{equation*}

For brevity, we drop the step-index $i$ and use $V^{\prime}$ for $V^{i+1}$.
Then, we expand $Q$ with second-order approximation with the coefficients  
\begin{subequations}\label{eq:Q expand}
\begin{align}
Q_{\mathbf{x}} &= l_\mathbf{x} + f_{\mathbf{x}}^T V^{\prime}_{\mathbf{x}} \label{eq:Qx} \\
Q_{\mathbf{u}} &= l_\mathbf{u} + f_{\mathbf{u}}^T V^{\prime}_{\mathbf{x}} \label{eq:Qu} \\
Q_{\mathbf{xx}} &= l_{\mathbf{xx}} + f_{\mathbf{x}}^T V^{\prime}_{\mathbf{xx}} f_{\mathbf{x}} + V^{\prime}_{\mathbf{x}} \cdot f_{\mathbf{xx}} \label{eq:Qxx} \\
Q_{\mathbf{uu}} &= l_{\mathbf{uu}} + f_{\mathbf{u}}^T V^{\prime}_{\mathbf{uu}} f_{\mathbf{u}} + V^{\prime}_{\mathbf{x}} \cdot f_{\mathbf{uu}} \label{eq:Quu} \\
Q_{\mathbf{ux}} &= l_{\mathbf{ux}} + f_{\mathbf{u}}^T V^{\prime}_{\mathbf{xx}} f_{\mathbf{x}} + V^{\prime}_{\mathbf{x}} \cdot f_{\mathbf{ux}} \label{eq:Qux}
\end{align}
\end{subequations}
where the subscripts denote differentiation. With this approximation, the optimal control modification $\mathbf{\optimal{\delta u}}$ is obtained as %Equation~\ref{eq:optimal u}
\begin{equation} \label{eq:optimal u}
    \mathbf{\optimal{\delta u}} = -Q_{\mathbf{uu}}^{-1} (Q_{\mathbf{u}} + Q_{\mathbf{ux}} \mathbf{\delta x}),
\end{equation}
and by ignoring the second order derivative of the dynamics (i.e. $f_{\mathbf{xx}}, f_{\mathbf{uu}}, f_{\mathbf{ux}}$), each coefficient can be recursively obtained with
\begin{subequations}~\label{eq:value relation}
\begin{align}
\Delta V(i) &= -\frac{1}{2}Q_\mathbf{u} Q_{\mathbf{uu}}^{-1} Q_{\mathbf{u}} \label{eq:deltaV} \\
V_{\mathbf{x}}(i) &= Q_{\mathbf{x}} - Q_{\mathbf{u}} Q_{\mathbf{uu}}^{-1} Q_{\mathbf{ux}} \label{eq:Vx} \\
V_{\mathbf{xx}}(i) &= Q_{\mathbf{xx}} - Q_{\mathbf{xu}} Q_{\mathbf{uu}}^{-1} Q_{\mathbf{ux}} \label{eq:Vxx}.
\end{align}
\end{subequations}

For more detailed derivations, we refer the readers to the previous work by \citet{mayne_differential_1973}.

%%%%%%%%%% Problem Formulation %%%%%%%%
\section{Problem Formulation} \label{sec:problem formulation}
In this section, we introduce essential notations to formulate the \emph{spatio-temporal motion retargeting} (STMR) problem.
In the latter part of this section, we effectively solve the proposed STMR problem by decoupling it into two stages: spatial motion retargeting (SMR) and temporal motion retargeting (TMR).
In essence, the SMR problem addresses only the kinematic aspects, while the TMR problem considers the dynamics of the target robotic system.

\subsection{Notations} \label{sec:notations}
Consider a robot with $\maxJ$ joints whose joint position is denoted as $\jp \in \mathbb{R}^\maxJ$.
The base position of the robot is denoted as $\pb \in \mathbb{R}^3$, and base orientation represented with quaternion is denoted as $\h \in \mathbb{H}$ where $\mathbb{H}$ is unit quaternion space.
The generalized coordinate of the robot is defined by stacking these values denoted as $\mathbf{q} = [\pb, \h, \jp ]$. Similarly, we write linear, angular, joint velocity as $\mathbf{v} \in \mathbb{R}^3, \mathbf{w} \in \mathbb{R}^3$, and $\mathbf{\dot{\jp}} \in \mathbb{R}^{\maxJ}$, respectively. The time derivative of generalized coordinate is accordingly defined as $\dot{\mathbf{q}} =[\mathbf{v}, \mathbf{w}, \mathbf{\dot{\jp}}]$ and the state of the robot is defined as $\mathbf{x}=[\mathbf{q}, \dot{\mathbf{q}}]^T$.

\begin{figure}[!t]
    \centering
    \captionsetup[subfloat]{labelfont=normal, font=normal}
    \subfloat[Temporal deformation]{
        \includegraphics[width=0.96\linewidth]{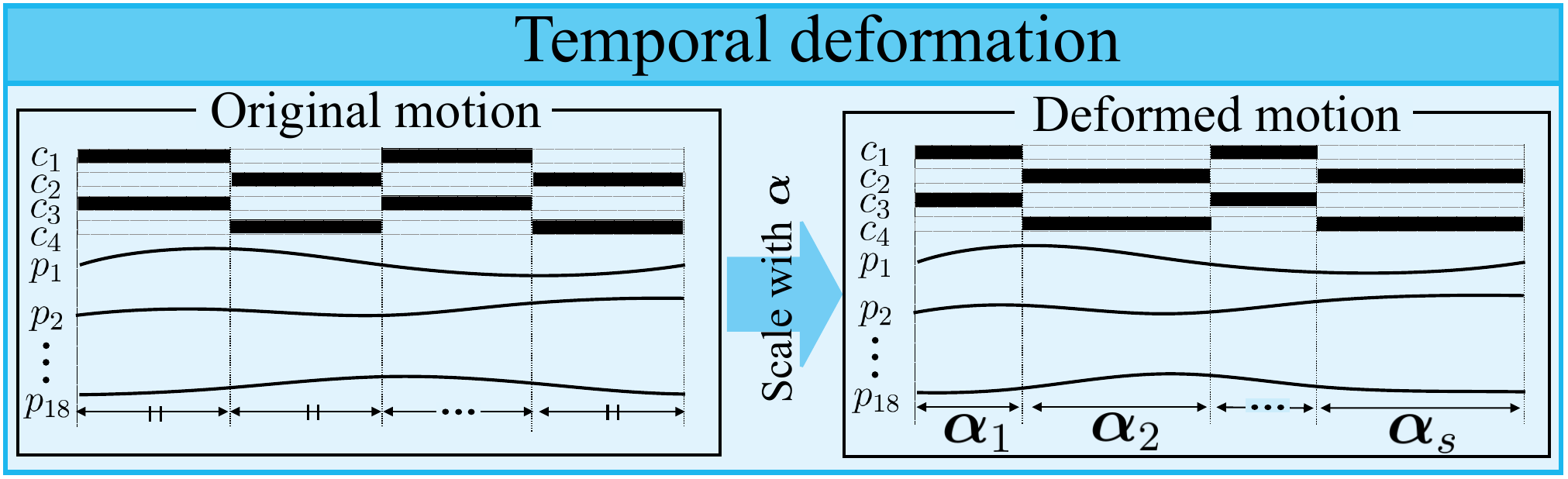}
        \label{fig:Temporal deformation}
    } \\
    \vspace{0.1cm}
    \subfloat[Temporal motion retargeting]{
        \includegraphics[width=0.94\linewidth]{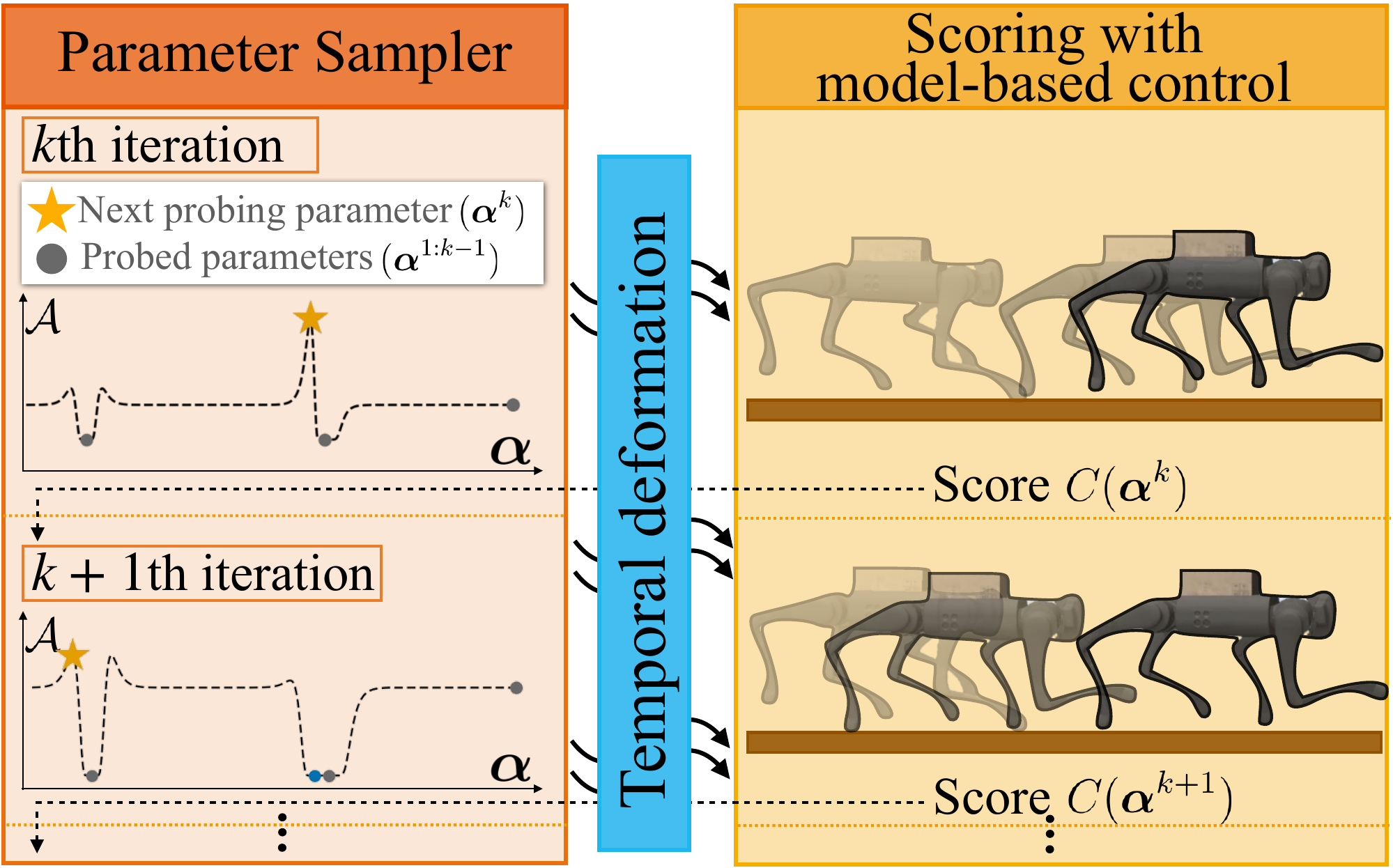}
        \label{fig:TMR overview}
    }
    \caption{The overview of temporal motion retargeting (TMR) is illustrated. (a) Temporal deformation involves splitting the motion into equal segments and scaling with temporal parameter $\alphabf$. (b) Using acquisition function $\mathcal{A}$, the temporal parameter $\alphabf$ is sampled for temporal deformation and scored by model-based control.}
    \label{fig:TMR method}
\end{figure}

We denote the keypoint positions as $\p \in \mathbb{R}^{N \times 3}$ and $\maxK$ is the number of keypoints. 
Specifically, we focus on $N=16$ keypoints \reviewPrev{consisting of four hips, thighs, knees, and feet.}
These values are specified for each frame, with the maximum frame index denoted by $\maxF$. 
For instance, the position of the $j$th keypoint in the $i$th frame is denoted as $\p_j^i$, where $j \in [1, 2, ..., \maxK]$ and $i \in [0, 1, 2, ..., \maxF]$.
Furthermore, we define foot index $\kappa(\cdot)$ to represent position of four feet as $\p_{\kappa(1)}, \p_{\kappa(2)}, \p_{\kappa(3)}, \p_{\kappa(4)}$. 

We write forward kinematics as $\text{FK}_j$, which maps $\mathbf{q}$ to the global position of the $j$th keypoint as $\p_{j} = \text{FK}_j(\q)$, and its jacobian matrix as $\mathbf{J}_j$, where $\mathbf{J}_j = \frac{\partial{(\indexK{\FK}{j}} (\mathbf{q}))}{\partial \mathbf{q}} \in \mathbb{R}^{(\maxJ + 6) \times 3}$.
We note that, in the subsequent chapter, we abuse the forward kinematic notation without the subscript as $\mathbf{p} = \FK(\x)$ to denote a mapping between the full state $\mathbf{x}$ and the concatenated keypoint positions $\p$, for brevity.

\subsection{Spatio-temporal motion retargeting} \label{sec:STMR problem}
% Overview goal of STMR
As the keypoint trajectory $\p$ is acquired from an arbitrary quadruped actor, it can be physically infeasible for the target robot to track.
Furthermore, the keypoint trajectory $\p$ may not contain the base movement required for whole-body imitation. 
To this end, we propose STMR, which regenerates the physically feasible whole-body motion by optimizing both spatial and temporal dimensions.

Let us define the temporal parameters as $\alphabf \in \mathbb{R}^{\maxSeg}$ that scale the keypoint motion along the time axis, which we refer to as \emph{temporal deformation}, as shown in \Cref{fig:Temporal deformation}. 
More specifically, we divide the source motion into $\maxSeg$ segments and temporally scale each segment by the factor corresponding to each component of $\alphabf$. 
We define this operation as the temporal deformation function $s_\alphabf(t)$, which maps control time to the corresponding time in the source motion. 
Through this operation, $T$ frames of keypoint trajectories correspond to $T_\alphabf$ control time steps.
Additionally, we define the linear interpolation function, $\LI(t; \Keymotion)$, which computes the keypoint positions at continuous time $t \in \mathbb{R}$ by linearly interpolating the keypoint trajectories. Therefore, $\LI(s_\alphabf(t_k); \Keymotion)$ gives the keypoint positions at $k$th control time step, which serves as control targets.

Then, we formulate an optimization problem as described in \Cref{eq:objective} where search for the optimal temporal parameters $\optimal{\alphabf}$, control sequence $\optimal{\mathbf{u}}^{0:T_{\optimal{\alphabf}}}$, and robot states $\optimal{\x}^{0:T_{\optimal{\alphabf}}}$.
\begin{gather} 
    \begin{aligned}
        \min_{{\mathbf{u}}^{0:\maxF_\alphabf},{\x}^{0:\maxF_\alphabf}, \alphabf \in \mathcal{I}} \quad & \frac{1}{2 \maxF_\alphabf} \sum_{k=0}^{\maxF_\alphabf} \|
        \FK(\x^k)- \LI\big(s_\alphabf(t_k);\Keymotion)\big)
        \|_Q^2 \\
        \text{s.t.} \quad & {\mathbf{x}}^{k+1} = f(\mathbf{x}^k, \mathbf{u}^k), \\
        & h(\mathbf{x}^k, \mathbf{u}^k) \leq 0, \\
        & g_{\text{eq}}(\mathbf{x}^k) = 0, \\
        & g_{\text{in}}(\mathbf{x}^k) \leq 0,
    \end{aligned}
    \label{eq:objective} \raisetag{60pt}
\end{gather}
where, $t_k$ is the control time at $k$th time step and $\mathcal{I}$ is the predefined bound for the temporal parameters $\alphabf$. 

In the STMR problem, the system is constrained by the dynamics $f$ and other physical constraints $h$, such as torque limit, Coulomb friction cone constraints, so on.
Additionally, we introduce foot constraints $g_\text{eq}$ and $g_\text{in}$ that prevent foot sliding and ground penetration, while enforcing identical contact timing to the original motion. The details of the foot constraints will be described in \Cref{sec:constraints}.

It is worth noting that the STMR problem involves transforming the target motion to make it feasible for the robot to imitate rather than simply tracking the motion.
Therefore, the retargeted motion should be constructed to preserve the semantic meaning of the original motion.
For that reason, we deform the motion in the bounded temporal regions $\mathcal{I}$ to preserve the overall expressiveness of the motion. 

\reviewPrev{Solving this optimization problem is challenging due to its complexity, which involves nonconvex objectives and constraints.}
To address this, we divide this problem into two subproblems: spatial motion retargeting (SMR) and temporal motion retargeting (TMR).
\reviewPrev{This decomposition simplifies the problem, even though it remains nonconvex, as detailed in the following sections.}

% Notably, due to the \reviewPrev{complexity of the problem, which involves nonconvex objectives and constraints, solving this optimization is challenging.}
% To address this, we divide this problem into two subproblems: spatial motion retargeting (SMR) and temporal motion retargeting (TMR).
% \reviewPrev{While both subproblems remain nonconvex even after decomposition, this approach simplifies the overall problem, with details presented in the subsequent sections.}

\subsection{Spatio-Temporal decoupling}\label{sec:two stage}
Due to the challenges mentioned earlier, we decompose the STMR problem into separate spatial and temporal components. 
With this approach, we sequentially determine mappings for each component through a two-stage optimization process.

In the first stage, the SMR stage, we focus on the kinematics of the motion. Starting from the STMR problem in \Cref{eq:objective}, we exclude the dynamics $f$ and temporal parameters $\alphabf$, concentrating on the kinematic feasibility of the position-level state $\q$ by enforcing the foot constraints $g$.
We search for kinematic feasible states $\qsmr$ by minimizing the objective function under the foot constraints $g$, as shown in \Cref{eq:SMR_objective_overview}. \reviewPrev{Specifically, foot constraints are applied to prevent foot sliding and penetration while ensuring identical contact timing. We refer the reader to \Cref{sec:constraints} for more details on these foot constraints.}

\begin{equation} \label{eq:SMR_objective_overview}
    \begin{aligned}
        \min_{{\q}^{0:\maxF}} \quad & \frac{1}{2 \maxF} \sum_{i=0}^{\maxF} \|
        \FK(\q^i)- \p^i
        \|_Q^2 \\
        \text{s.t.} \quad &  g_{\text{eq}}(\q^i) = 0, \\
        & g_{\text{in}}(\q^i) \leq 0
    \end{aligned}
\end{equation}

Following this, we compute the trajectory of keypoints corresponding to $\qsmr$ as $\bar{\mathbf{p}} = \FK(\qsmr)$. 
We then perform temporal deformation with $s_\alphabf$ on the newly obtained keypoint trajectory.
Finally, we solve TMR problem, presented in \Cref{eq:TMR_objective_fn}, to search for the optimal temporal parameters $\optimal{\alphabf}$, control sequence $\optimal{\mathbf{u}^{0:T_{\optimal{\alphabf}}}}$ and resulting states $\optimal{\mathbf{x}}^{0:T_{\optimal{\alphabf}}}$:
\begin{gather} 
    \begin{aligned} 
        \min_{{\mathbf{u}}^{0:\maxF_\alphabf},{\x}^{0:\maxF_\alphabf}, \alphabf \in \mathcal{I}} & \quad  \frac{1}{2 \maxF_\alphabf} \sum_{k=0}^{T_\alphabf} \|
        \FK(\x^k)- \LI\big(s_\alphabf(t_k);\bar{\mathbf{p}}^{0:T})\big)
        \|^2_Q\\
        \text{s.t.} \quad & {\mathbf{x}}^{i+1} = f(\mathbf{x}^k, \mathbf{u}^k), \\
        & h(\mathbf{x}^k, \mathbf{u}^k) \leq 0,  \\
        & g_{\text{eq}}(\mathbf{x}^k) = 0, \\
        & g_{\text{in}}(\mathbf{x}^k) \leq 0
    \end{aligned} 
    \label{eq:TMR_objective_fn} \raisetag{60pt}
\end{gather}

Note that the TMR problem encompasses a finite-horizon optimal control problem (OCP), which finds the optimal control sequence to track the given reference states under dynamics.
By ignoring the temporal parameter $\alphabf$, the TMR problem reduces to OCP with the goal of tracking the reference keypoint trajectory $\mathbf{p}^{0:T}$.
From this insight, we reformulate this problem as a nested optimization problem to reduce the complexity:
\begin{equation*}
\label{eq:nested}
\min_{\alphabf \in \mathcal{I}} 
\left(
    \begin{aligned} 
        \min_{{\mathbf{u}}^{0:\maxF_\alphabf},{\x}^{0:\maxF_\alphabf}} & \,  \frac{1}{2 \maxF_\alphabf} \sum_{k=0}^{T_\alphabf} \|
        \FK(\x^k)- \LI\big(s_\alphabf(t_k);\bar{\mathbf{p}}^{0:T})\big)
        \|^2_Q\\
        \text{s.t.} \quad & {\mathbf{x}}^{i+1} = f(\mathbf{x}^k, \mathbf{u}^k), \\
        & h(\mathbf{x}^k, \mathbf{u}^k) \leq 0,  \\
        & g_{\text{eq}}(\mathbf{x}^k) = 0, \\
        & g_{\text{in}}(\mathbf{x}^k) \leq 0
    \end{aligned} 
\right)
\end{equation*}
As illustrated in \Cref{fig:TMR overview}, we iteratively search for the optimal set of $\optimal{\alphabf}$, $\optimal{\x}^{0:T_{\optimal\alpha}}$ and $\optimal{\mathbf{u}}^{0:T_{\optimal\alpha}}$.
For each iteration of the outer loop, we begin with a given value of $\alphabf$ and solve the inner optimization for the $\optimal{\x}^{0:T_\alphabf}$ and $\optimal{\mathbf{u}}^{0:T_\alphabf}$. In the next iteration, we update $\alphabf$ and repeat the process until a minimum is reached. The resulting robot state $\optimal{\x}^{0:T_{\optimal\alpha}}$ serves as the kino-dynamically retarged motion.
More details on TMR are discussed in \Cref{sec:TMR}.

\reviewPrev{Furthermore, it is worth noting that we solve this internal optimal control problem using the off-the-shelf simulator~\citep{todorov2012mujoco} using full dynamics model $f$ instead of the reduced model.
This approach makes introducing a new target robot straightforward, enabling motion retargeting to arbitrary quadruped robots.
However, solving optimal control with full dynamics model $f$ can be challenging because of high-dimensional state space. 
In this paper, we utilize DDP from Section~\ref{sec:ddp}, which is known to be effective in such scenarios~\citep{howell_predictive_2022}.
}

%%%%%%%%% Methods %%%%%%%%%%%%%
\begin{figure*}[!t]
    \centering
    \captionsetup[subfloat]{labelfont=normal, font=normal}
    \subfloat[{Baseline Method}]{
        \includegraphics[width=0.98\linewidth]{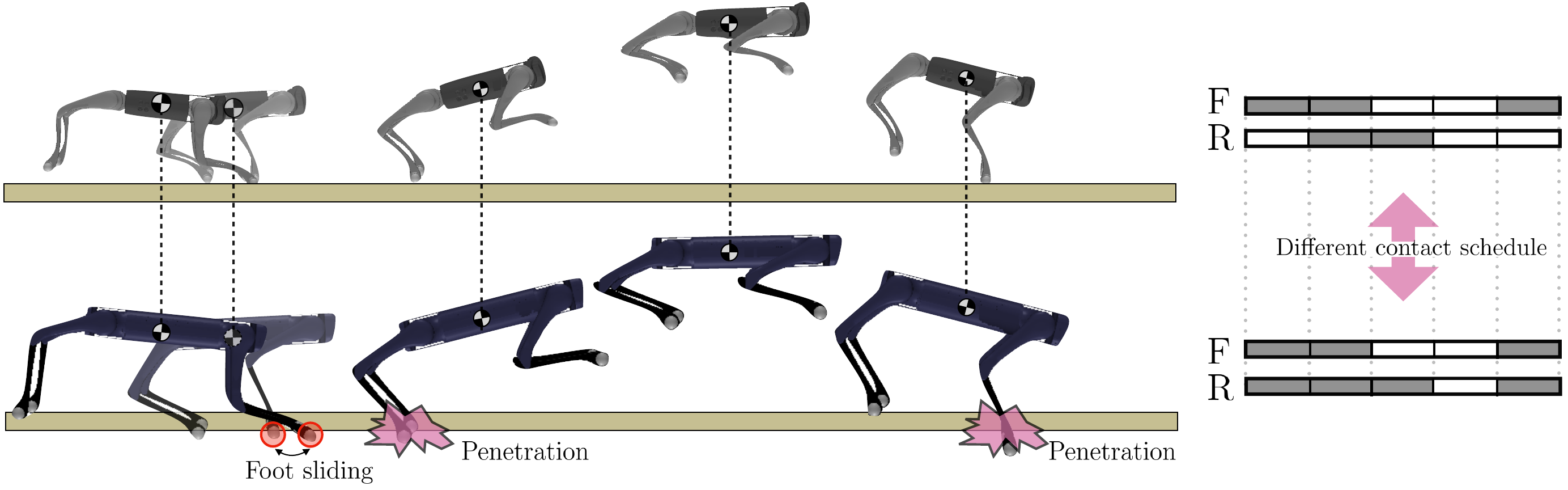}
        \label{fig:problem}
    } \\
    \vspace{0.4 cm}
    \subfloat[{Spatial motion retargeting}]{
        \includegraphics[width=0.98\linewidth]{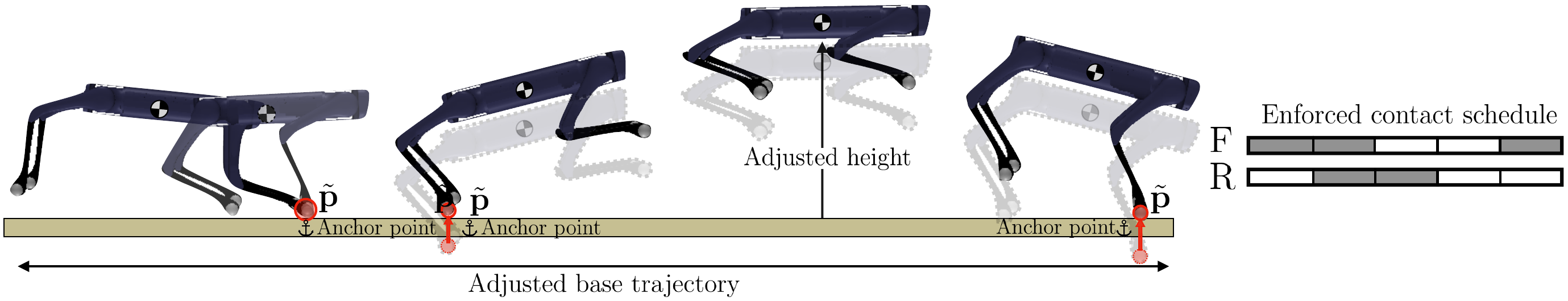}
        \label{fig:SMR_detail}
    } 
    \caption{Illustration of baseline method and spatial motion retargeting (SMR). (a) The baseline method (i.e., Unit Vector Method) can lead to kinematic artifacts such as foot sliding, foot penetration, and mismatched contact timing. (B) On the other hand, SMR generates kinematically feasible motion by appropriately anchoring the foot position as $\tilde{\p}$.}
    \label{fig:SMR_method}
\end{figure*}

\section{Spatial Motion Retargeting}\label{sec:SMR}
\begin{algorithm} [!t]
\caption{Spatial Motion Retargeting}
\small
\setstretch{1.05}
\begin{algorithmic}[1] 
\State ${\overhorizon{\mathbf{q}}{0}} \leftarrow \overhorizon{\quv}{0}$
\State $\tilde{\p} \leftarrow \textsc{ProjectGround}(\FK(\mathbf{q}))$
\State ${\dquv} \leftarrow \textsc{Differentiate}(\quv^1, \quv^0)$
\For{$i=1$ \textbf{to} $N$} 
    \State $\dquv \leftarrow \textsc{Differentiate}(\overhorizon{\quv}{i}, \overhorizon{\quv}{i+1}, \Delta t)$
    \If {\textbf{not} \textsc{Any($\con^i$})} \Comment{Flight phase}
        \If {\textsc{Any($\con^{i-1}$})}
            \State $\vexit \leftarrow \textsc{Polyfit}(\textsc{BasePosition}(\q^{0:i}))$
        \EndIf
        \State $\textsc{BaseVelocity}(\dquv) \leftarrow \vexit$
        \State $\vexit \leftarrow \vexit + \mathbf{g} \Delta t$
    \EndIf
    \State $\qref \leftarrow \textsc{Integrate}({\overhorizon{\mathbf{q}}{i-1}},\dquv, \Delta t)$
    \Repeat
        \State $\mathbf{J} \leftarrow \textsc{GetJacobian}(\mathbf{q})$
        \State $\dot{\mathbf{q}} \leftarrow \textsc{SolveQP}(\mathbf{J}, \q, \qref, \con, K)$ \Comment{\Cref{eq:SMR}}
        \State $\overhorizon{\mathbf{q}}{i} \leftarrow \overhorizon{\mathbf{q}}{i} + \eta \dot{\mathbf{q}}$
    \Until {$\dot{\mathbf{q}} < \dot{\mathbf{q}}_\text{thres}$}
    \For {$j=1$ \textbf{to} $4$}
        \State $\p_{\kappa(j)} = \indexK{\FK(\q)}{\kappa(j)}$
        \If {$\p_{\kappa(j)^z} < \textsc{HeightMap}(\p_{\kappa(j)}^{xy})$}
            \State $\indexK{\overhorizon{\con}{i}}{j} \leftarrow \text{True}$
        \EndIf
        \If {$\indexK{\overhorizon{\con}{i}}{j}$ \textbf{and} \textbf{not} $\indexK{\overhorizon{\con}{i-1}}{j}$} \Comment{At the new contact segment}
            \State $\indexK{\tilde{\p}}{j}  \leftarrow \textsc{ProjectGround}(\p_{\kappa(j)})$
        \EndIf
    \EndFor
\EndFor
\end{algorithmic} \label{alg:SMR}
\end{algorithm}

In this section, we provide more details on spatial motion retargeting (SMR) that retargets motion at the kinematic level.
One simple way for kinematic motion retargeting is the Unit Vector Method (UVM), as outlined in \Cref{sec:unit vector}, that maintains the directional unit vector between adjacent keypoints. 
However, this approach may introduce undesired artifacts such as foot sliding or foot penetration.
For instance, \Cref{fig:problem} illustrates that transferring a walking trajectory by a short-legged robot to a long-legged robot leads to unnatural foot sliding and penetration.
Additionally, the UVM cannot generate a base trajectory beyond direct transfer, making it unable to adjust the base trajectory to fit the robot's kinematic configuration. More critically, this naive transfer using the UVM can alter contact schedules, failing to preserve the semantic meaning of the original motion.

On the other hand, our SMR, illustrated in \Cref{fig:SMR_detail}, eliminates foot sliding and foot penetration, adjusts base trajectory and heights according to the target robot's kinematics, and preserves contact schedules of the source motion.
As briefly mentioned in \Cref{sec:problem formulation}, we introduce the foot constraints, formulated as two constraints: the \emph{contact preservation constraint}, which ensures that the contact schedule of the source and retargeted motion remain identical without any foot penetration, and the \emph{foot locking constraint}, which prevents foot sliding during contact phases. 
By enforcing these constraints, SMR aims to obtain a refined generalized coordinate trajectory of robot $\Qtar$ that mimics the source motion.
The overall algorithm of SMR is summarized in \Cref{alg:SMR}.

\reviewPrev{
Let us elaborate more on the intuition of SMR in whole-body motion reconstruction from baseless motion.
A key idea is minimizing the positional difference of the locally defined baseless motion while optimizing the whole-body motion. 
However, without additional constraints, this often results in a trivial solution where the robot flounders in the air. 
Introducing foot constraints helps mitigate this issue by effectively regularizing the resulting motions.
Intuitively speaking, we are anchoring the robot's feet while adjusting its joint angles to induce base movement. 
The whole-body motion is reconstructed by repeatedly detaching and re-anchoring the feet according to a given contact schedule while adjusting each joint.
In essence, \emph{foot constraints are not just an outcome of SMR but a fundamental component in reconstructing whole-body motion.}
}

In practice, SMR can be used to retarget motion in two scenarios: when the base trajectory is provided and when it is not. 
In the first case, SMR efficiently suppresses kinematic artifacts when given whole-body motion, as evaluated in \Cref{sec: kinematic artifacts}. 
In the second case, SMR reconstructs whole-body motion by incorporating contact schedules, even when only the local movements of keypoints are known, which we evaluate in \Cref{sec:eval recon}.
\reviewPrev{Moreover, we demonstrate that the base trajectory generation is powerful enough to adapt the motion to the given terrain, as shown in \Cref{sec:backflip_realworld}.}

\subsection{Foot Constraints} \label{sec:constraints}
Let $j$th foot of a robot be in a contact phase. 
We introduce the anchor position of $j$th foot as $\indexK{\tilde{\p}}{j}$, which is a projection of $\text{FK}_j(\q)$ to the ground.
During the contact phase, the height of the foot should match the elevation of the projection point, and in a swing phase, it should be positioned above this point. 
This condition can be expressed as 
\begin{equation}\label{eq:C1}
    \begin{cases}
    \indexK{\FK}{j}(\q)^{z} = \indexK{\tilde{\p}}{j}  ^{z}, & \text{if } c_j\\
    \indexK{\FK}{j}(\q)^{z} > \indexK{\tilde{\p}}{j}  ^{z}, & \text{else},
  \end{cases}
\end{equation}
where $z$ represents the height component of the position vector, and $c_j$ is a contact boolean for $j$th foot.

Additionally, we fix the x and y coordinates of the foot position to the anchor point during the contact phase to eliminate foot sliding as follows: %whose position $\indexK{\tilde{\p}}{j}$ is updated for the new contact segment.
\begin{equation}\label{eq:C2}
    \indexK{\FK}{j}(\q)^{xy} = \indexK{\tilde{\p}}{j}  ^{xy} \  \text{ if } c_j.
\end{equation}

\reviewPrev{we combine \Cref{eq:C1} and \Cref{eq:C2}, and relax them as follows, which we refer to as $g(\q)$:}
\begin{equation} \label{eq:constraints}
    c_j(\indexK{\FK}{j}(\q) - \indexK{\tilde{\p}}{j}) = 0
\end{equation}

\subsection{Objective function} \label{sec:SMR_mimic}
%The optimization problem presented in \Cref{eq:SMR_objective_overview} is not straight forward to solve as it is non-convex problem.
On top of the relaxation in the foot constraints, we make the objective function in \Cref{eq:SMR_objective_overview} convex. %near the solution of unconstrained version of this problem.
We start by incorporating the relaxed constraints $g$ obtained in \Cref{eq:constraints} into \Cref{eq:SMR_objective_overview} and search for the optimal generalized coordinate sequence $\bar{\q}^{0:T}$ that minimize the positional distance to keypoints as described in \Cref{eq:SMR_q_objective}:
\begin{equation} \label{eq:SMR_q_objective}
    \begin{aligned}
        \bar{\q}^{0:T} = \arg \min_{{\q}^{0:\maxF}} \quad & \frac{1}{2 \maxF} \sum_{i=0}^{\maxF} \|\FK(\q^i)- \p^i\|^2_Q \\
        \text{s.t.} \quad &  g(\q^i) = 0
    \end{aligned}
\end{equation}

Without the foot constraints $g$, \Cref{eq:SMR_q_objective} becomes a typical unconstrained inverse kinematics (IK) problem. 
Instead of solving \Cref{eq:SMR_q_objective} directly, we solve the unconstrained IK problem to obtain the generalized coordinate solution $\quv^{0:T}$ and then compute its time derivative $\dquv^{0:T}$ by finite difference. 
In the subsequent steps, the time derivative of IK solutions serves as the velocity target for the final optimization problem. 
This allows us to transform the \Cref{eq:SMR_q_objective} to a velocity-level problem, which is more straightforward to solve.

More specifically, we build the reference $\qref^i$, which is obtained by time-integrating the current coordinate $\q^i$ with $\dquv^i$.
Additionally, we write scaled error between reference coordinates $\qref^i$ and current coordinate $\q^i$ as $\dqref^i = {K_q}(\qref^i - \mathbf{q}^i)$, where ${K_q}$ is a tunable parameter.
Finally, we form the optimization problem mimicking the reference $\qref^i$ for a single frame, subject to the linearized foot constraints, as follows:
\begin{equation} \label{eq:SMR}
\begin{split}
    \dot{\bar{\mathbf{q}}}^i = &\arg \min_{\dot{\mathbf{q}}^i} \, \frac{1}{2} 
    \|\dot{\mathbf{q}}^i - \dqref^i\|^2_Q \\
    &\text{s.t. } \, \mathbf{J}^i_j \dot{\mathbf{q}}^i = c_j {K_p} (\indexK{\tilde{\p}}{j} - \indexK{\FK}{j}(\q^i)) \quad j \in [1,2,3,4].
\end{split}
\end{equation}

We solve this problem sequentially, starting from the initial frame to the last. This approach has the additional benefit of simplifying the determination of the foot anchor point $\tilde{\p}$. 
Together with the linearized foot constraints, this approach ensures that the system effectively satisfies the foot constraints.

The final form of the problem is a convex optimization problem, which allows us to use off-the-shelf solvers. In this work, we utilize the ADMM~\citep{admm} method implemented by \citet{osqp}.

\subsection{Handling flight phases}
As SMR heavily relies on contact schedules, it is crucial to handle scenarios where all of a robot's feet are in swing phases (i.e., flight phases). In such cases, we assume the robot's base follows a ballistic trajectory. % We calculate the exit velocity $\vexit$ at the beginning of the flight phase and update the position based on gravitational acceleration $\mathbf{g}$.

As described in \Cref{alg:SMR}, we fit a polynomial function to the history of base trajectories when a flight phase is detected. Specifically, choose the degree of the polynomial based on the current time step clipped by the maximum horizon of $h$.
Then, we calculate the exit velocity $\vexit$ by taking the derivative of the polynomial function. 
We update the velocity by integrating gravitational acceleration $\mathbf{g}$, allowing the robot to follow the ballistic trajectory until the contact schedule is set or new contact occurs between the robot and the terrain.

In addition, note that such a ballistic trajectory is set for the base trajectory of reference motion $\qref$.
Since we solve the optimization process to determine the final kinematic posture $\q$, the whole-body motion will be adjusted to correspond to the ballistic base trajectory of the reference motion $\qref$.
% As a result, we obtain natural motion, such as a soft landing with knees bending according to the base trajectory.

\section{Temporal Motion Retargeting} \label{sec:TMR}
% topic: we use BO and MBOC
In this section, we elaborate on temporal motion retargeting (TMR), which generates dynamically feasible motions for the target robot by determining the temporal parameters $\alphabf$ and control sequence $\U$.
The main challenge of TMR lies in jointly optimizing both temporal parameters $\alphabf$ and control sequence $\U$. 
As outlined in \Cref{sec:STMR problem}, we approach this as a nested optimization problem, where we iteratively search for the optimal temporal parameters $\optimal{\alphabf}$ using Bayesian Optimization (BO)~\citep{NIPS2012_05311655}. %to improve sample efficiency.
\reviewPrev{In detail, TMR involves the repetition of three processes, as shown in \Cref{fig:TMR overview}}: parameter sampling, temporal deformation, and scoring.

For a warm start, we begin by randomly sampling the temporal parameter within the interval $\mathcal{I} = [\alphabf_{\text{min}}, \alphabf_{\text{max}}]$. 
Subsequently, we perform temporal deformation, denoted as $s = s_\alphabf(t)$, which divides the time frame into $\maxSeg$ equal intervals and scales the motion according to the temporal parameters $\alphabf$, as illustrated in \Cref{fig:Temporal deformation}.

Subsequently, we employ MBOC to track the deformed motion and evaluate the result with the scoring function denoted as $C(\cdot)$. 
The key intuition is that appropriately deforming the motion makes it easier for MBOC to track, leading to improved tracking performance.
Specifically, we design the scoring function to evaluate not only keypoint tracking and contact matching but also to regularize extreme values of the temporal parameters, enhancing motion imitation performance in practice.
\Cref{eq:cost} presents the metric, $C(\alphabf)$, for evaluating the tracking performance which incorporates measurements of contact differences using Intersection over Union (IoU), base positional error (L1 distance $d_b$), and base orientation error (L1 distance of Euler angles $d_E$), where $\pb$ and $\h$ represent base position and orientation, respectively.
\begin{equation} \label{eq:cost}
    C(\alphabf) = -d_b(\mathbf{p}_\text{b}, \bar{\mathbf{p}}_\text{b}) - d_E(\mathbf{h}, \barbf{\h}_\text{b}) + w_c \text{IoU}(\con, \barbf{\con})
\end{equation}

Denoting the temporal parameter at the $k$th iteration as $\alphabf^k$ and its corresponding score as $C^k$, the next probing point, $\alphabf^{k+1}$, is sampled based on the previous probing points $(\alphabf^{1:k}, C^{1:k})$.
To achieve this, we first fit a surrogate function $s$ using Gaussian process regression:
\begin{equation} \label{eq:GP}
    \begin{gathered}
        s(\alphabf) \sim \mathcal{GP}(s_\mu(\alphabf), s_\sigma(\alphabf)), \\
        s_\mu = \kernel_{*}^{T} \kernel^{-1} C^{1:k}, \quad s_\sigma = \kernel_{**} -  \kernel_{*}^{T} \kernel^{-1}\kernel_*,
    \end{gathered}
\end{equation}
where $\kernel$, $\kernel_*$ and $\kernel_{**}$ are defined as follows, using a Matern kernel $\kernel(\cdot, \cdot)$:
\begin{equation}
    \begin{gathered}
        \kernel = \kernel(\alphabf^{1:k},\alphabf^{1:k}), \\
        \kernel_* = \kernel(\alphabf, \alphabf^{1:k}), \qquad 
        \kernel_{**} = \kernel(\alphabf, \alphabf).    
    \end{gathered}
\end{equation}

We then construct an acquisition function, $\mathcal{A}$, using Expected Improvement (EI)~\citep{NIPS2012_05311655}, where $\hat{\alphabf}$ represents the best parameter found in the history, $\phi$ denotes the Gaussian distribution, $\Phi$ is the cumulative distribution function (CDF) of $\phi$, and $\xi$ controls the degree of exploration:
\begin{equation} \label{eq:acqusition}
    \begin{gathered}
        \mathcal{A} = \Delta s \Phi(\Delta s/ s_\sigma) + s_\sigma \phi(\Delta s/ s_\sigma),\\
        \Delta s = s_\mu(\alphabf) - s_\mu(\hat{\alphabf}) + \xi, \quad \hat{\alphabf} = \argmax_{\alphabf \in \{\alphabf_1, \alphabf_2, \ldots, \alphabf_k\}} C(\alphabf).
    \end{gathered}
\end{equation}
Finally, we sample the next probing parameter, $\alphabf^{k+1}$, by maximizing the acquisition function, $\alphabf^{k+1} = \arg \max_\alphabf \ac(\alphabf)$. 
This process is repeated until a value of $\alphabf$ converges.

\begin{figure} [!t]
    \centering
    \captionsetup{labelfont=normal, font=normal}
    \includegraphics[width=0.99\linewidth]{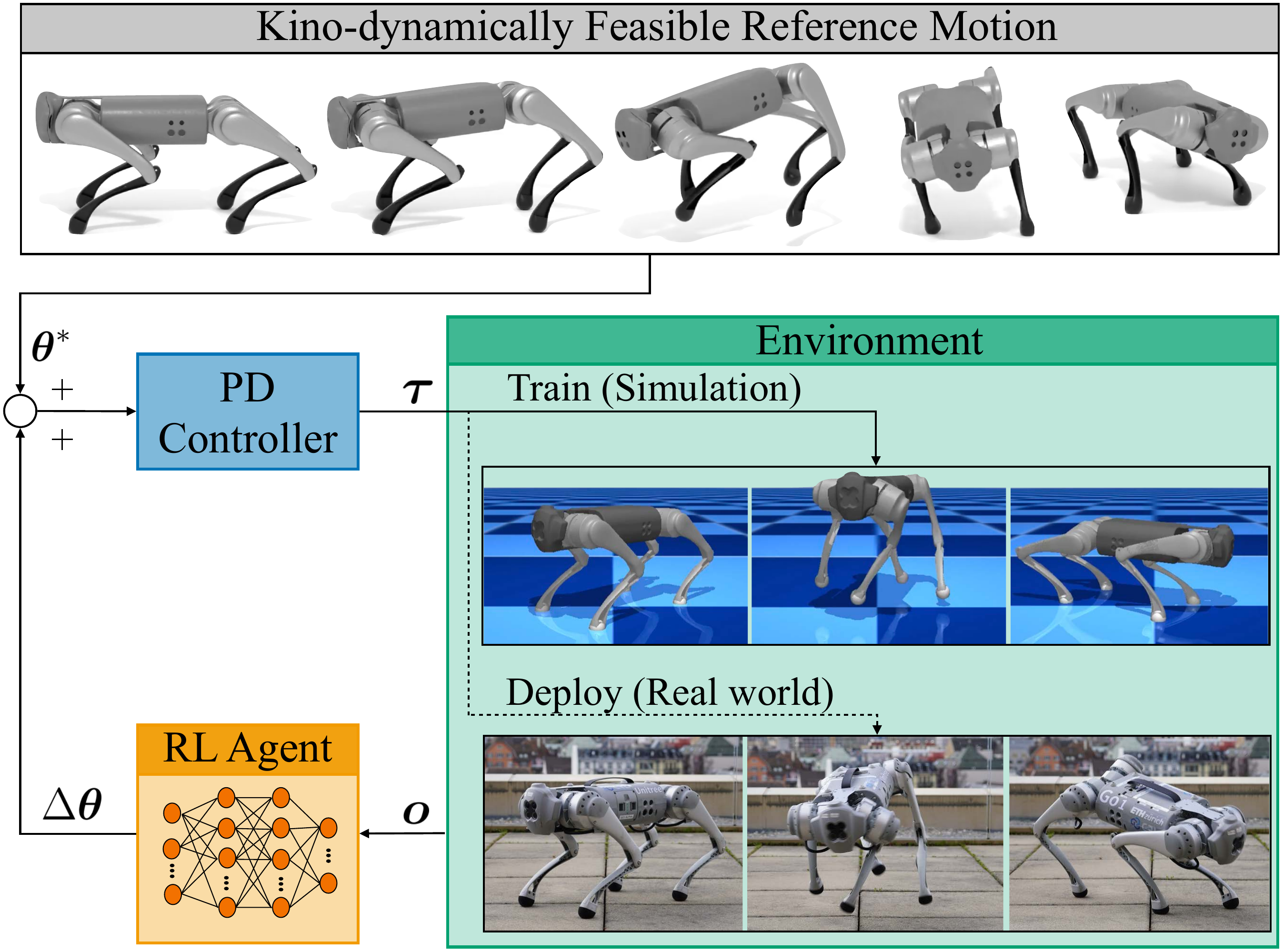}
    \caption{Control policy with residual learning}
    \label{fig:policy_learning}
\end{figure}

We note that any dynamics model capable of producing whole-body motion can be used for MBOC. 
In our experiments, we used a full-body model implemented through the MuJoCo engine \citep{todorov2012mujoco, howell_predictive_2022}, as MuJoCo provides the derivative information required for DDP described in \Cref{sec:ddp}. 
Since we use a full-body model, it only requires a Universal Robot Description File (URDF) or its extensions. This allows us to avoid the tedious modeling process typically required for reduced models.

\section{Residual policy learning} \label{sec:policy learning}

The optimal control sequence identified in the STMR stage can successfully produce the retargeted motion in simulation with open-loop control execution. However, to deploy the motions robustly in the real world, a feedback policy is necessary to overcome uncertainties and model mismatches.
Therefore, we adopt residual policy learning~\citep{residual_learning} to guide the feedback policy learning process.
The residual policy denoted as $\pi$, computes a closed-loop control signal that is added to a base control $\optimal{\X}$ obtained through STMR.
As illustrated in \Cref{fig:policy_learning}, the joint values of feasible motion $\optimal{\jp}$ are obtained directly from indexing the feasible motion $\optimal{\X}$ where the residual control policy outputs the joint correction denoted as $\Delta{\jp}$.
Both terms are fed to the PD controller to yield a motor torque command as $\tau=K_p(\optimal{\jp}+\Delta{\jp}-\jp) - K_d\dot{\jp}$, where $K_p$ and $K_d$ are the proportional gains and derivative gains, respectively, and $\jp$ and $\dot{\jp}$ are the current robot's joint angles and velocities.

The residual policy is trained with RL, where the rewards at frame $i$ are defined as a summation of tracking measures for joint position $\indexK{\overhorizon{\jp}{i}}{1:\maxJ}$, base position $\overhorizon{\qb}{i}$, base orientation $\overhorizon{\h}{i}$, the position of keypoints $\indexK{\overhorizon{\p}{i}}{1:\maxK}$:
\begin{equation*} \label{eq:reward}
\begin{split}
    \overhorizon{r_t}{i} = 
        w_q \exp[\beta_\qj \|
            \overhorizon{\indexK{\bar{\bm{\theta}}}{1:\maxJ}}{i} - 
            \indexK{\overhorizon{{\bm{\theta}}}{i}}{{1:\maxJ}}\|]
        + w_\text{b} \exp[\beta_\text{b} \|
            \overhorizon{{\bar{\mathbf{p}_\text{b}}}}{i} -
            \overhorizon{{{\mathbf{p}_\text{b}}}}{i} \|] \\
        + w_\h \exp[\beta_\h \|
            \overhorizon{\bar{\h}}{i}  \ominus 
            \overhorizon{\h}{i} \|]
        + w_\p \exp[\beta_\p \|
            \overhorizon{\bar{\p}}{i}_{1:\maxK} -
            \overhorizon{\p}{i}_{1:\maxK} \|].
\end{split}    
\end{equation*}
The values of the hyperparameters used in our experiments are presented in \Cref{tab:train config}.

The observation space, denoted as ${\bm{o}}$, is defined in \Cref{eq:observation} which consists of projected gravity $\mathbf{g}_\text{proj}$, joint positions $\jp$, joint velocity $\dot{\jp}$, last deployed torque $\bm{\tau}$, phase variable $\phi$, and base height $\mathbf{p}_\text{b}^z$:
\begin{equation} \label{eq:observation}
    \bm{o} = [\mathbf{g}_\text{proj}, \jp, \dot{\jp}, \mathbf{\tau}, \phi, \mathbf{p}_\text{b}^z].
\end{equation}
% We note that the observation can also include base linear velocity and angular velocity, if they are available from the base state estimator. % Real-world Experiment 으로 이동

To facilitate learning dynamic motions involving flying phases, we adopt the \emph{reference state initialization} scheme from {DeepMimic}~\citep{peng_deepmimic_2018}, which determines the initial state of each episode through uniform sampling from the reference motion.

\begin{table}[!t]
    \captionsetup{font=small}
    \caption{List of the hyper-parameters used for RL training.}
    \centering
    \footnotesize
    \setstretch{1.13}
    \begin{tabular}{|c|c|} 
        \hline
         Hidden dimensions (Actor) & [512, 256, 128] \\
        \hline
         Hidden dimensions (Critic) & [512, 256, 128] \\
        \hline
         Activation function & Exponential Linear Unit  (ELU) \\
        \hline
         Weight coefficient for Entropy term & 0.01 \\
        \hline
         Learning rate & 1.0e-3 \\
        \hline
         Discount factor & 0.99 \\
         \hline
        Optimizer & ADAM~\citep{kingma2014adam} \\
        \hline
        Number of policy iterations & 10,000 \\
        \hline
        Joint reward ($w_q$, $\beta_\qj$) & (3.0, 2.0)\\
        \hline
        Base position reward ($w_\text{b}$, $\beta_\text{b}$) & (3.0, 5.0)\\
        \hline
        Base orientation reward ($w_\h$, $\beta_\h$) & (3.0, 1.0)\\
        \hline
        Keypoints reward($w_\p$, $\beta_\p$) & (30.0, 10.0)\\
        \hline 
        $K_p$, $K_d$ (A1) & (30.0, 1.0) \\
        \hline 
        $K_p$, $K_d$ (Go1) & (30.0, 1.0) \\
        \hline 
        $K_p$, $K_d$ (AlienGo) & (50.0, 1.0) \\
        \hline 
        $K_p$, $K_d$ (B2) & (200.0, 10.0) \\
        \hline
    \end{tabular}
    \label{tab:train config}
\end{table}

%%%%%%%%%%% Results %%%%%%%%%%%
\section{Simulational Experiments}
We conducted a series of simulation experiments to verify the efficacy of our approach.
In the first experiment, we evaluated the tracking performance of the final motion execution on simulated robots and demonstrated the superiority of our method by comparing it to baseline imitation learning (IL) methodologies.
In the second experiment, we validated that the motion transferred by our method is free of foot sliding and preserves the contact schedules.
We then demonstrate that our method can reconstruct the base movements from baseless motions and quantify the recovery rate.
\reviewPrev{Finally, we highlight the importance of temporal optimization using the BackFlip motion, a highly dynamic movement that includes a significant flight phase.}

% Average values are listed with standard deviations in parentheses.
\begin{figure*} [!t]
    \centering
    \captionsetup{labelfont=normal, font=normal}
    \includegraphics[width=0.99\linewidth]{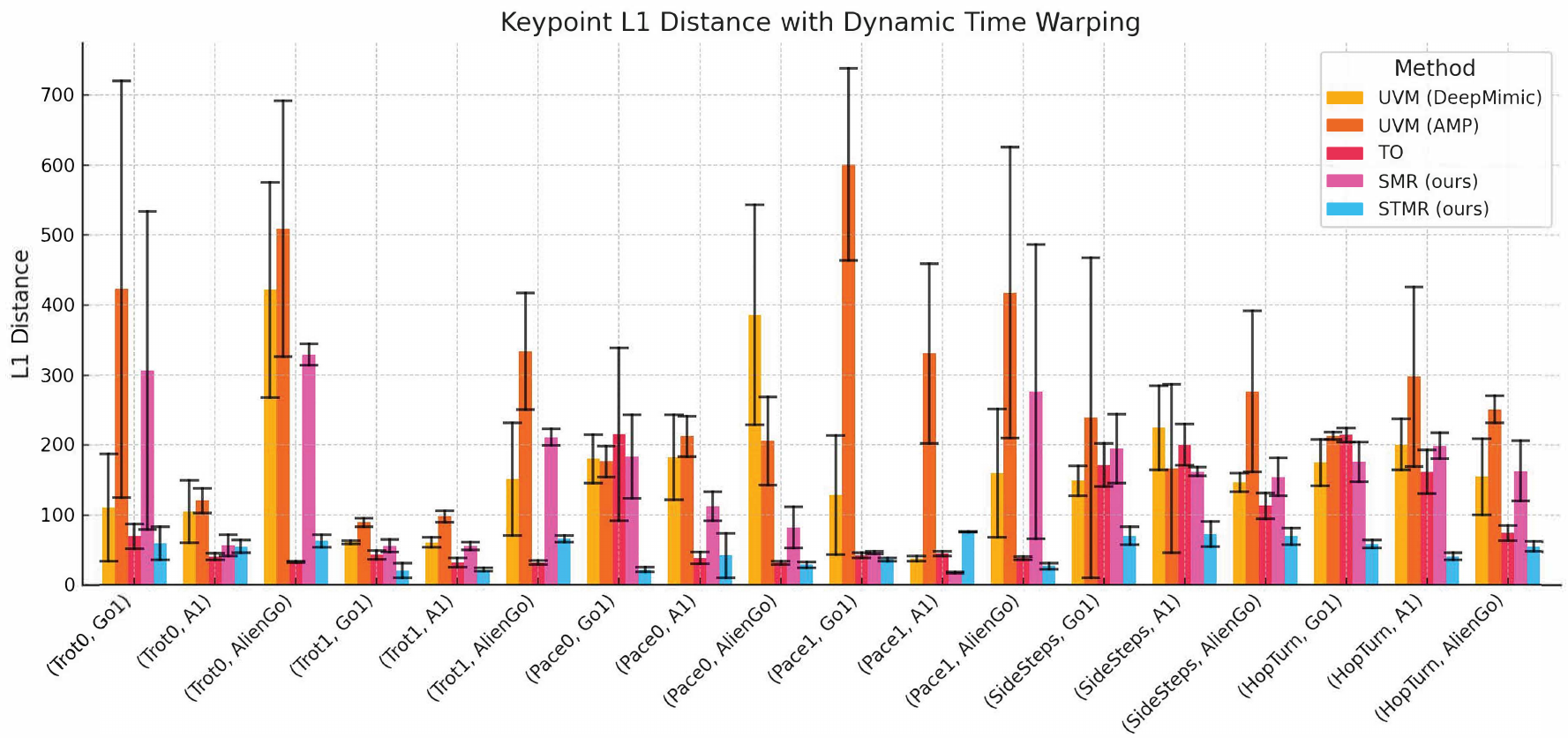}
    \caption{\reviewPrev{The tracking performance for each robot and motion is illustrated in terms of the mean and standard deviation.}
    }
    \label{fig:barplot}
    \vspace{0.2cm}
\end{figure*}

\subsection{Training details}
All RL control policies used in our experiments were trained with simulational data generated by Isaac Gym~\citep{makoviychuk2021isaac}.
We employed Proximal Policy Optimization (PPO)~\citep{schulman2017proximal} with 10,000 iterations, equivalent to approximately 50 million data samples requiring one hour of training with NVIDIA RTX A6000 GPU.
For each experiment, we trained five policies with five different random seeds and reported the mean and standard deviation of the metrics.
More details on training configuration are summarized in \Cref{tab:train config}.

\subsection{Details of baseline methods}
\label{sec:baselines}
We selected three other motion retargeting baselines, including downstream IL methods, to evaluate how the motions retargeted in both space and time domains lead to more successful IL.

\begin{table*}[!t]
    \captionsetup{font=small}
    \caption{Normalized Keypoint L1 distance with Dynamic time warping (\%).}
    \centering
    \footnotesize
    \setstretch{1.13}
    \begin{tabular}{|p{1.2 cm}|p{1.2cm}||p{2.0cm}|p{2.0cm}|p{2.0cm}|p{2.0cm}|p{2.0cm}|}
    \hline
    \multirow{2}{*}{\begin{tabular}{@{}l@{}}Motion\end{tabular}} &
    \multirow{2}{*}{\begin{tabular}{@{}l@{}}Robot\end{tabular}} &
    \multirow{2}{*}{\begin{tabular}{@{}l@{}}\shortstack[l]{\reviewPrev{UVM}\\ \reviewPrev{(DeepMimic)}}~\citep{peng_deepmimic_2018}\end{tabular}} &
    \multirow{2}{*}{\begin{tabular}{@{}l@{}}\shortstack[l]{\reviewPrev{UVM}\\ \reviewPrev{(AMP)}}~\citep{peng_amp_2021}\end{tabular}} &
    \multirow{2}{*}{\begin{tabular}{@{}l@{}}\reviewPrev{TO}~\citep{fuchioka_opt-mimic_2023}\end{tabular}} &
    \multirow{2}{*}{\begin{tabular}{@{}l@{}}\reviewPrev{SMR (ours)}\end{tabular}} &
    \multirow{2}{*}{\begin{tabular}{@{}l@{}}STMR (ours)\end{tabular}} \\
    & & & & & &\\
    \hline
    \hline
    \multirow{3}{*}{SideSteps} & Go1 & $4.0 (0.6)$ & $6.4 (6.1)$ & $3.5 (0.6)$ & $\reviewPrev{5.3(1.3)}$& $\textbf{1.9} (0.3)$\\
    \cline{2-7}
     & A1 & $7.2 (1.9)$ & $5.3 (3.8)$ & $5.0 (0.7)$ & $\reviewPrev{5.4 (0.2)}$ & $\textbf{2.7} (0.7)$\\
    \cline{2-7}
     & AlienGo & $3.7 (0.3)$ & $6.9 (2.9)$ & $2.7 (0.4)$ & $\reviewPrev{3.7 (0.6)} $& $\textbf{1.7} (0.3)$\\
    \hline
    \multirow{3}{*}{HopTurn} & Go1 & $6.5 (1.2)$ & $7.9 (0.2)$ & $6.1 (0.3)$ & $\reviewPrev{6.4(1.0)}$ & $\textbf{2.1} (0.2)$\\
    \cline{2-7}
     & A1 & $8.2 (1.5)$ & $12.2 (5.2)$ & $4.4 (0.8)$ & $\reviewPrev{8.2 (0.7)}$ & $\textbf{1.6} (0.2)$\\
    \cline{2-7}
     & AlienGo & $5.3 (1.9)$ & $8.5 (0.7)$ & $2.1 (0.3)$ & $\reviewPrev{5.2 (1.4)}$ & $\textbf{1.7} (0.2)$\\
    \hline
    \end{tabular}
    \label{tab:normalized tracking result}
\end{table*}

\subsubsection{\reviewPrev{UVM (DeepMimic)}} \label{subsec:deepmimic}
\reviewPrev{
We aim to show that motion retargeted by the proposed method leads to better policy learning than the Unit Vector Method (UVM).
In detail, we employ the UVM and direct base position transfer for motion retargeting and train the control policy to imitate the retargeted motions. 
The downstream control policy learning is the same as STMR, including the reward function from \Cref{eq:reward}.
Therefore, the main difference between this baseline method and STMR is using the UVM for motion retargeting. 
This retargeting schema is essentially equivalent to DeepMimic~\citep{peng_deepmimic_2018}, which employs motion retargeting through the direct transfer of joint angles, except that we do not scale the base position manually for each robot.
}

\subsubsection{Adverserial motion prior (AMP)}
\reviewPrev{
We use the \emph{AMP}~\citep{escontrela_adversarial_2022} approach as a second baseline method to evaluate whether motion retargeting plays a critical role in successful motion imitation or if improved control policy learning alone is sufficient. 
By leveraging a reward derived from a learned discriminator, the AMP approach can partially circumvent the need for dynamically feasible reference motions, as demonstrated by \citet{li2023learning}. 
Consequently, we apply the motion obtained through the UVM in conjunction with AMP's control policy learning and compare the results to those of the proposed method.
}

For AMP, we trained a discriminator $D$ to classify whether the transition between the current state $s$ and the next state $s^\prime$ is generated by the agent or from a reference motion.
The output from the discriminator $D$ contributes an additional reward term with the corresponding weight $w_{\text{ad}}$ as
\begin{equation} \label{eq:amp_reward}
    r = r_t + w_{\text{ad}} \log{D(s,s^{\prime})}.
\end{equation}
As a side note, we extended the training iterations to 25,000 for AMP policies to achieve convergence.

\subsubsection{TO}
\reviewPrev{
Our method is comparable to \emph{OptMimic}\citep{fuchioka_opt-mimic_2023}, as both approaches use model-based optimal control to obtain a dense description of dynamic motion. 
Specifically, OptMimic employs a single rigid body model\citep{single_rigid_body} and contact-implicit trajectory optimization with a non-convex optimizer~\citep{wachter2006implementation} to refine the reference motion. 
\reviewPrev{In more detail, they incorporate constraints to avoid foot penetration, enforce friction cones, and complementary conditions for contacts.}
Lastly, this baseline follows the same control policy learning framework as described in \Cref{subsec:deepmimic}, making a fair comparison.

A key distinction between our proposed method and this baseline lies in the scope of optimization: while OptMimic focuses on spatial dimensions, our approach incorporates optimization in both spatial and temporal dimensions. 
Therefore, we demonstrate the impact of temporal optimization by comparing the performance of our method against this baseline.
}

\subsubsection{STMR (Ours)}
As mentioned in \Cref{sec:SMR}, STMR can retarget motions with and without the global base pose trajectory.
In the simulation experiments, we used the base pose trajectories in \Cref{sec:evaluate tracking} and \Cref{sec: kinematic artifacts}, while we reconstructed the base pose trajectory solely from local movement in \Cref{sec:eval recon}.

We set the number of time segments to $\maxSeg=1$, as the tested motions for \Cref{sec:evaluate tracking} are relatively short, as shown in \Cref{table:motion_details}.
This is possible because Differential Dynamic Programming (DDP), which serves as the internal process for temporal motion retargeting (TMR), can provide more fine-grained temporal adjustments.
\reviewPrev{Note that in \Cref{sec:videos}, we set the number of time segments to $\maxSeg=3$ as the reference motions include different phases of motions, such as slow walking led by quick turn.
Lastly, the bounds for the temporal parameters are set as $\log_2{\alphabf} \in [-1, 1]$, meaning the motion can be scaled by a factor of up to two in either direction.
}

\begin{table}
    \captionsetup{font=small}
    \caption{
    The source and duration of the motion dataset are illustrated. Mcp refers to motion capture data, and Anim refers to handmade animation data. The durations are given in seconds. 
    }
    \centering
    \footnotesize
    \setstretch{1.12}
    \begin{tabular}{|c|c|c|c|c|c|c|}
    \hline
    & Trot0  & Trot1 & Pace0 & Pace1 & SideSteps & HopTurn \\
    \hline
    Source & Mcp & Mcp & Mcp & Mcp & Anim & Anim\\
    \hline
    Duration & 1.65 & 1.65 & 1.95 & 2.50 & 14.50 & 9.10 \\
    \hline
    \end{tabular}
    \label{table:motion_details}
\end{table}

\subsection{Evaluating motion tracking performance} \label{sec:evaluate tracking}

\reviewPrev{
We compare the tracking performance of baseline methods and the proposed method to evaluate how appropriate the retargeted motions are.
The intuition of measuring tracking performance is that if retarget motions are dynamically feasible for the target robot, it will be straightforward for the controller to follow them, resulting in a smaller tracking error.
Similarly, better tracking performance indicates a reduced likelihood of imitation failure, as a policy that fails to reproduce a source motion on a robot results in significantly lower tracking performance.
}
We utilize Dynamic time warping (DTW)~\citep{senin2008dynamic} to measure distance irrelevant to temporal deformation, avoiding the risk of under-evaluating baseline methods.
Details on the motion retargeting method used for each IL approach are presented in \Cref{sec:baselines}. 

\reviewPrev{We used motion clips collected from real dogs from \citet{zhang2018mode} and quadruped motions crafted manually by animators from \citet{peng2020learning}. 
As these motions are collected from diverse sources, we demonstrate the versatility of our method.}
Regarding the complexity of these motions, they can be ranked in ascending order as follows: Trot, Pace, SideSteps, and HopTurn. 
Trot motion is comparatively more straightforward to replicate than Pace as it involves cross-arranging two feet for enhanced stability.
Both Pace and SideSteps involve statically unstable postures. 
However, SideSteps poses a greater challenge due to the need to balance against lateral momentum.
Lastly, HopTurn is the most complex, as it requires jumping and a mid-air maneuver to turn.
In terms of the target robot, we employ three different sizes of quadrupedal robots (Unitree A1, Unitree Go1, and Unitree AlienGo).
In addition, we elongated the duration of each motion with a scale of two to highlight the impact of temporal deformation. 
More details on each motion are summarized in \Cref{table:motion_details}.

\reviewPrev{\Cref{fig:barplot} illustrates tracking performance in terms of mean and standard deviation over five random seeds.
}
STMR exhibits exceptional performance across all six motions, achieving an average tracking error of $48.7$ $\MM$. In comparison, the average errors for the three baseline methods—UVM (DeepMimic), UVM (AMP), TO, SMR —are $168.3$ $\MM$, $275.3$ $\MM$, $88.4$ $\MM$, and $154.2$ $\MM$, respectively. 
Thus, our method's improvement in tracking error corresponds to 71.1\%, 82.3\%, 44.9\%, and 68.4\% reductions for each baseline method, corresponding to an average improvement of 66.7\%.
\reviewPrev{Moreover, the result shows that STMR can perform the two most challenging motions with flight phases (i.e., SideSteps and HopTurn) whereas other baseline methods fail\footnote{The footage of each comparison experiment can be found in the attached video.}. 
}
We also report the normalized keypoint tracking error (L1 distance) expressed as a percentage of the total trajectory length (measured in meters), computed using dynamic time warping. Bold values indicate the best performance. Values in parentheses denote standard deviations.

\reviewPrev{
The TMR process requires significantly more computation than SMR.
Therefore, it is essential to evaluate how each component contributes to performance.
Therefore, we also conduct ablation study of STMR by evaluating only the SMR stage.
As shown in \Cref{fig:barplot}, TMR enhances the overall dynamic feasibility of the motions.
In particular, we highlight that SMR fails for HopTurn and SideSteps, suggesting that TMR plays a crucial role in the flight phase by adjusting motion in the temporal dimension.
}

\begin{table}[!t]
    \captionsetup{font=small}
    \caption{
    Evaluation of foot sliding and contact preservation.
    }
    \centering
    \footnotesize
    \setstretch{1.12}
    \setlength{\tabcolsep}{2pt}
    \begin{tabular}{|c|c|c|c|c|c|c|c|}
    \hline
    {\multirow{2}{*}{\begin{tabular}{@{}l@{}}Robot   \end{tabular}}} & 
    {\multirow{2}{*}{\begin{tabular}{@{}l@{}}Motion\end{tabular}}} & 
    \multicolumn{3}{c|}{Foot slide \reviewPrev{($\mathrm{mm}$)}$\downarrow$} &
    \multicolumn{3}{c|}{IoU $\uparrow$} \\
    \cline{3-8}
     & & UVM~\citep{sjchoi_iros_natual} &\reviewPrev{TO~\citep{fuchioka_opt-mimic_2023}}& Ours & UVM~\citep{sjchoi_iros_natual} &\reviewPrev{TO~\citep{fuchioka_opt-mimic_2023}}& Ours \\
    \hline
    \hline
    Go1 & Trot0 & 110.19 & \reviewPrev{51.72} &\textbf{0.15} & 0.46 & \reviewPrev{0.40} & \textbf{1.00} \\
    & Trot1 & 72.93 & \reviewPrev{82.71} &\textbf{0.09} & 0.48 & \reviewPrev{0.38} & \textbf{1.00} \\
    & Pace0 & 88.90 & \reviewPrev{205.46} &\textbf{0.14} & 0.47 & \reviewPrev{0.62} & \textbf{1.00} \\
    & Pace1 & 61.33 & \reviewPrev{73.71} &\textbf{0.09} & 0.53 & \reviewPrev{0.89} & \textbf{0.99} \\
    & SideSteps & 34.74 & \reviewPrev{6.12} &\textbf{0.03} & 0.60 & \reviewPrev{0.95} & \textbf{1.00} \\
    & HopTurn & 33.35 & \reviewPrev{11.87} &\textbf{0.05} & 0.59 & \reviewPrev{0.82} & \textbf{1.00} \\
    \hline
    A1 & Trot0 & 101.39 & \reviewPrev{118.31} & \textbf{0.15} & 0.44 & \reviewPrev{0.47} & \textbf{1.00} \\
    & Trot1 & 86.29 & \reviewPrev{135.82} & \textbf{0.12} & 0.50 & \reviewPrev{0.39} & \textbf{1.00} \\
    & Pace0 & 83.29 & \reviewPrev{68.54} & \textbf{0.12} & 0.47 & \reviewPrev{0.75} & \textbf{1.00} \\
    & Pace1 & 63.58 & \reviewPrev{100.06} & \textbf{0.11} & 0.53 & \reviewPrev{0.82} & \textbf{1.00} \\
    & SideSteps & 41.37 & \reviewPrev{13.00} & \textbf{0.04} & 0.58 & \reviewPrev{0.84} & \textbf{1.00} \\
    & HopTurn & 37.19 & \reviewPrev{12.01} & \textbf{0.05} & 0.49 & \reviewPrev{0.79} & \textbf{1.00} \\
    \hline
    AlienGo & Trot0 & 147.01 & \reviewPrev{65.27} & \textbf{0.07} & 0.46 & \reviewPrev{0.48} & \textbf{1.00} \\
    & Trot1 & 112.22 & \reviewPrev{141.84} & \textbf{2.89} & 0.54 & \reviewPrev{0.40} & \textbf{0.98} \\
    & Pace0 & 114.51 & \reviewPrev{24.38} & \textbf{0.10} & 0.49 & \reviewPrev{0.76} & \textbf{1.00} \\
    & Pace1 & 71.88 & \reviewPrev{59.56} & \textbf{0.09} & 0.53 & 
    \reviewPrev{0.85} & \textbf{1.00} \\
    & SideSteps & 26.34 & \reviewPrev{0.32} & \textbf{0.04} & 0.50 & \reviewPrev{0.84} & \textbf{1.00} \\
    & HopTurn & 44.11 & \reviewPrev{14.44} & \textbf{1.83} & 0.58 & \reviewPrev{0.79} & \textbf{1.00} \\
    \hline
    \end{tabular}
    \label{table:footslide_iou}
\end{table}

\subsection{Evaluating foot constraints enforcement} \label{sec: kinematic artifacts}
SMR enforces foot constraints to generate kinematically feasible motions that preserve the original contact schedules and avoid foot sliding. 
Therefore, we quantitatively evaluate foot sliding and contact preservation using the same six motions from the previous \Cref{sec:evaluate tracking}. 
We measure foot sliding by calculating the L1 distance of position between the beginning and end of the contact segment, where continuous contact longer than $0.5$ seconds is marked as the contact segment.
Additionally, we measure the contact preservation through the Intersection over Union (IoU) between the contact schedules of the original and retargeted motion. 
An IoU of $1.0$ between two motions signifies identical contact schedules, indicating successful contact preservation, whereas an IoU of $0.0$ indicates completely divergent contact schedules.

To evaluate foot slide regularization and contact preservation, we compare against the UVM as the baseline method, where the results are summarized in \Cref{table:footslide_iou}. 
The proposed method shows an average foot sliding of $0.34$$\MM$, whereas that of the baseline method is $73.92$$\MM$ across six motions and three robots. 
Furthermore, the proposed method also shows significant improvement in contact preservation, with an average IoU of $0.998$ compared to $0.513$ for the baseline method. 
Given that the foot sliding error and the IoU achieve near-optimal values, the results strongly suggest that the motions generated by our method effectively eliminate foot sliding and accurately preserve contact schedules.

\reviewPrev{
Moreover, our method outperforms TO in both foot sliding prevention and contact preservation.
This is because foot sliding constraints are not explicitly enforced in the trajectory optimization framework, as doing so would distort the motion's semantic meaning. 
For instance, when retargeting a casual walking motion from a large robot to a smaller one, enforcing foot sliding constraints locks the feet to the ground, forcing unnatural leg spreading to match the global motion.
This alteration fundamentally changes the motion, making it no longer resemble casual walking.
Additionally, TO employs contact-implicit trajectory optimization, allowing contact booleans to change for improved tracking. 
However, this flexibility leads to deviations from the original contact schedules, resulting in a lower IoU.
In summary, STMR outperforms TO by generating a base trajectory tailored to each robot, enabling direct incorporation of constraints without conflicting with the global motion.
}

\begin{table}[!t]
    \captionsetup{font=small}
    \caption{
    Evaluation of spatial motion retargeting with and without the base motion. 
    }
    \centering
    \footnotesize
    \setstretch{1.12}
    \begin{tabular}{|@{\hspace{3pt}}c@{\hspace{3pt}}|@{\hspace{3pt}}c@{\hspace{3pt}}|c|c|c|}
    \hline
    {\multirow{4}{*}{\begin{tabular}{@{}l@{}}Robot   \end{tabular}}} & 
    {\multirow{4}{*}{\begin{tabular}{@{}l@{}}Motion\end{tabular}}} & 
    \multicolumn{3}{c|}{Distance Traveled \reviewPrev{by} } \\
    & & \multicolumn{3}{c|}{\reviewPrev{Retargeted Motions}$(\mathrm{m})$} \\
    \cline{3-5}
    & & \reviewPrev{With Base}& \reviewPrev{Without Base} & Recovery \\
    & & \reviewPrev{Motion} & \reviewPrev{Motion} & Rate $(\%)$\\
    \hline
    \hline
    Go1 & Trot $(0.5 \mathrm{m/s})$ & 0.97& 0.76& 78.35\\
    \reviewPrev{(Source)}& Trot $(1.0 \mathrm{m/s})$     & 1.93& 1.49& 77.20\\
    & Trot $(1.5 \mathrm{m/s})$     & 2.70& 1.93& 71.48\\
    & Pace $(1.0 \mathrm{m/s})$     & 1.91& 1.47& 76.96\\
    & Bound $(1.0 \mathrm{m/s})$    & 1.92& 1.42& 73.96\\
    & \reviewPrev{SideSteps} & \reviewPrev{0.243} & \reviewPrev{0.125} & \reviewPrev{51.44} \\
    \hline
    A1 & Trot $(0.5 \mathrm{m/s})$  & 0.93& 0.72& 77.42\\
    \reviewPrev{(Target)}& Trot $(1.0 \mathrm{m/s})$     & 1.84& 1.40& 76.09\\
    & Trot $(1.5 \mathrm{m/s})$     & 2.59& 1.82& 70.27\\
    & Pace $(1.0 \mathrm{m/s})$     & 1.80& 1.36& 75.56\\
    & Bound $(1.0 \mathrm{m/s})$    & 1.83& 1.33& 72.68\\
    & \reviewPrev{SideSteps} & \reviewPrev{0.234} & \reviewPrev{0.167} & \reviewPrev{71.80} \\
    \hline
    AlienGo & Trot $(0.5 \mathrm{m/s})$ & 1.12& 0.91& 81.25\\
    \reviewPrev{(Target)}& Trot $(1.0 \mathrm{m/s})$         & 2.18& 1.74& 79.82\\
    & Trot $(1.5 \mathrm{m/s})$         & 3.05& 2.28& 74.75\\
    & Pace $(1.0 \mathrm{m/s})$         & 2.16& 1.72& 79.63\\
    & Bound $(1.0 \mathrm{m/s})$        & 2.16& 1.67& 76.85\\
    & \reviewPrev{SideSteps} & \reviewPrev{0.254} & \reviewPrev{0.171} & \reviewPrev{67.32} \\
    \hline
    \end{tabular}
    \label{table:recon}
\end{table}

\subsection{Evaluating reconstruction of whole-body motion} \label{sec:eval recon}
Our STMR method is capable of generating the whole-body motion from the baseless keypoint trajectories as described in \Cref{fig:SMR}.
In this experiment, we quantitatively evaluate this capability by removing base motions from a set of source motions, reconstructing them, and measuring the recovery rate.
Moreover, we highlight that this reconstruction process generates base trajectories tailored to the kinematics of each of the three different robots, allowing for the efficient transfer of motions while overcoming morphological differences~\footnote{\reviewPrev{The footage of motion reconstruction can be found in the attached video.}}.

\begin{figure}[!t]
    \centering
    \captionsetup[subfloat]{labelfont=normal, font=normal}
    \subfloat[Trot]{
        \includegraphics[width=0.98\linewidth]{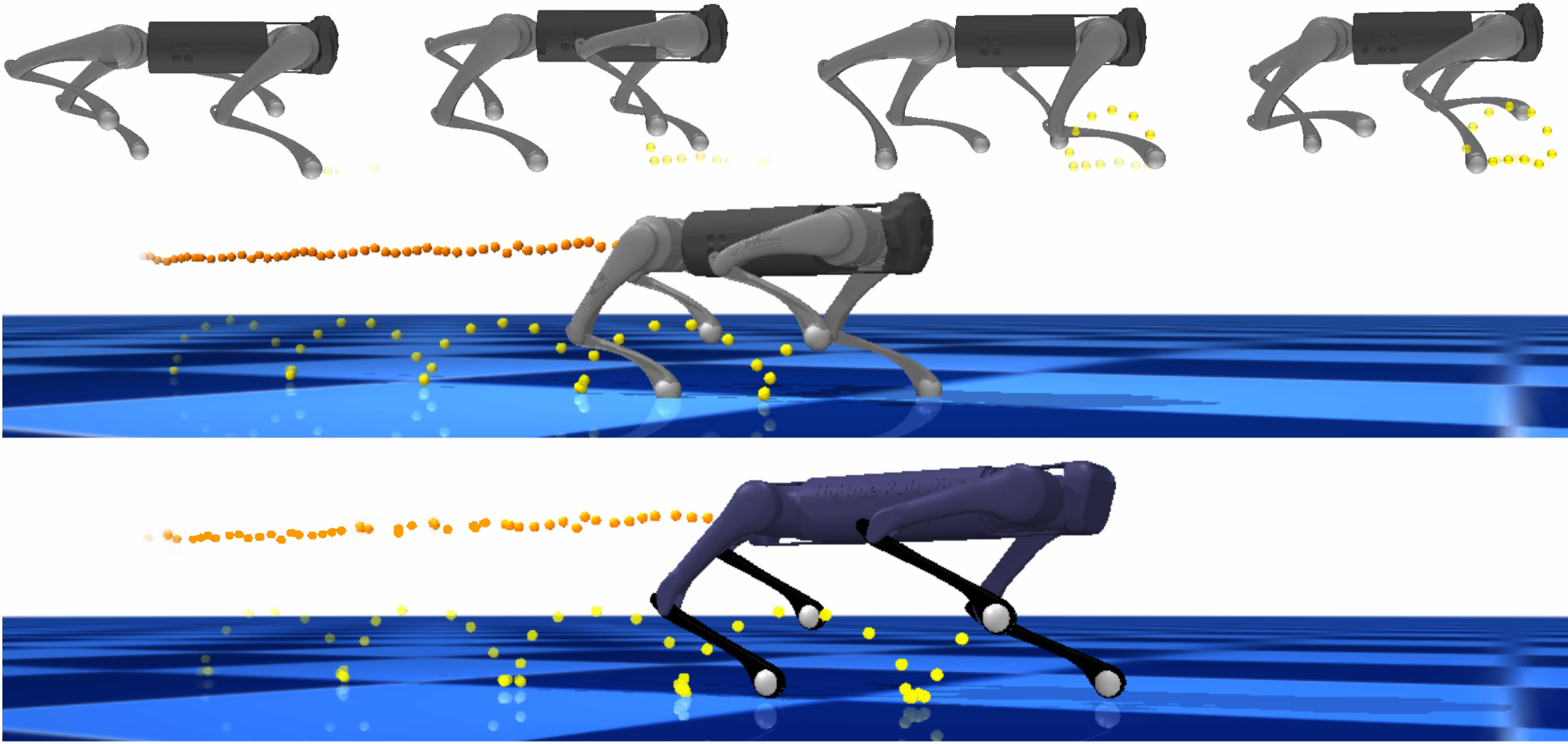}
        \label{fig:Recon trot}
    } \\
    \vspace{0.2cm}
    \subfloat[Pace]{
        \includegraphics[width=0.98\linewidth]{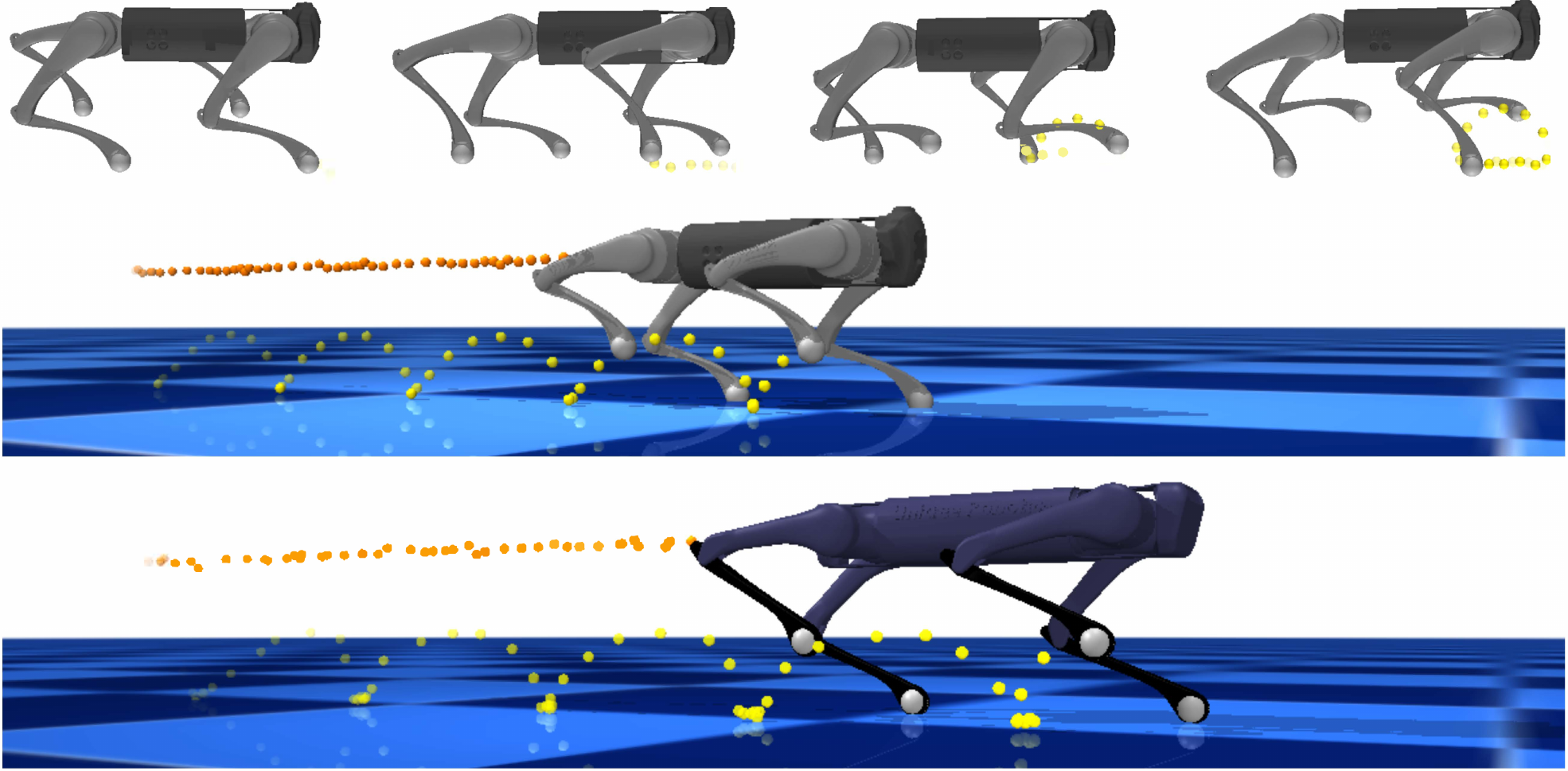}
        \label{fig:Recon pace}
    } \\
    \vspace{0.2cm}
    \subfloat[Bound]{
        \includegraphics[width=0.98\linewidth]{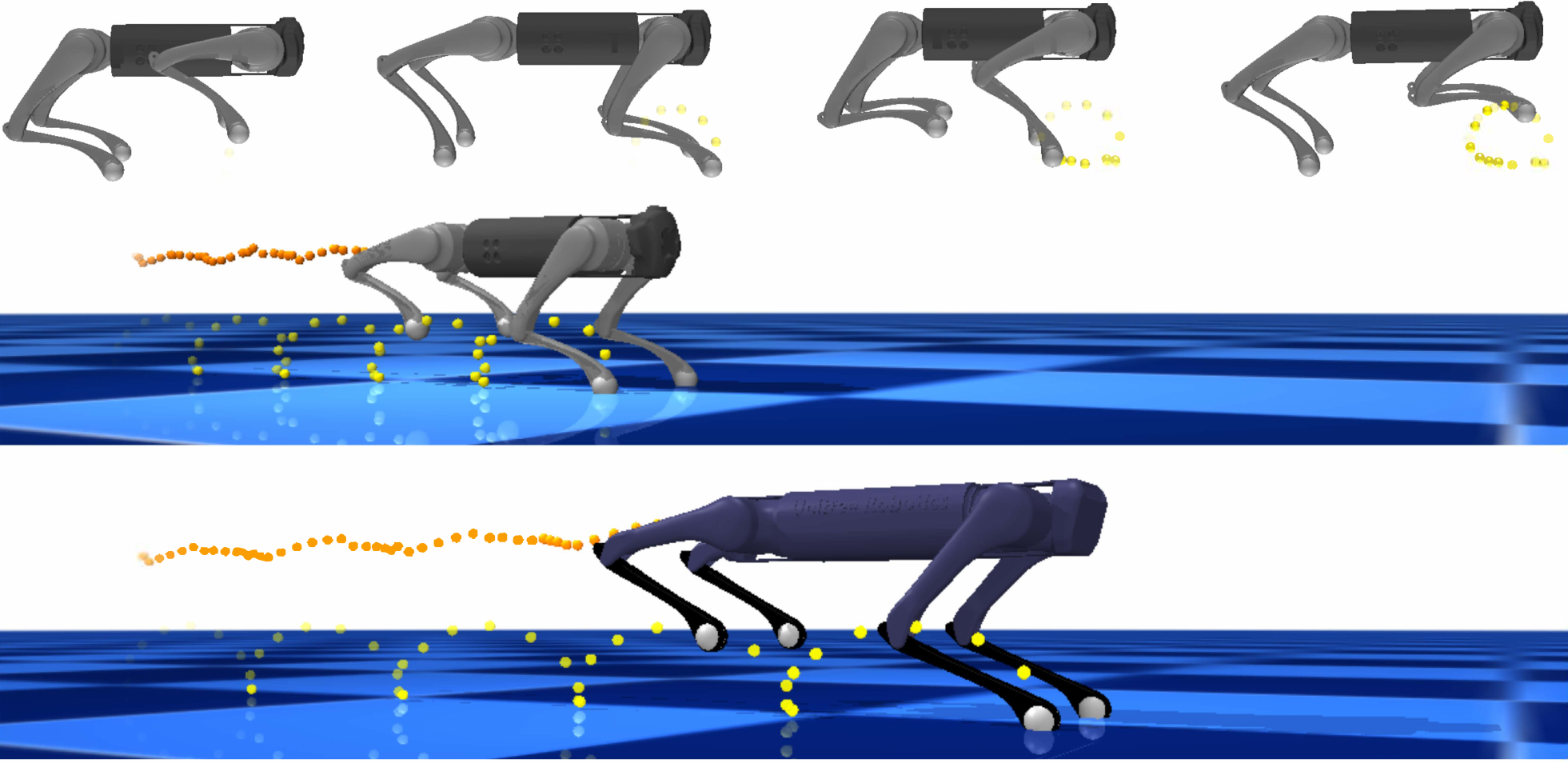}
        \label{fig:Recon bound}    
    }
    \caption{\reviewPrev{The differences between original motion and retargeted motion are shown for}: (a) Trot ($1.0 \mathrm{m/s}$), (b) Pace ($1.0 \mathrm{m/s}$), and (c) Bound ($1.0 \mathrm{m/s}$).
    \reviewPrev{The baseless motion of the small robot (Go1, Above) can reconstructed as whole-body motion (Go1, Middle). Similarly, it can be retargeted to the large robot (AlienGo, Below) without base motion. 
    The figure also shows that SMR can generate naturally elongated base trajectories for robots of different sizes.}
    The orange and yellow dots indicate base and foot trajectory, respectively.
    }
    \label{fig:Recon}
    \vspace{-0.5em}
\end{figure}

\begin{figure}[!h]
    \centering
    \captionsetup[subfloat]{labelfont=normal, font=normal}
    \subfloat[\reviewPrev{Go1 BackFlip}]{
        \includegraphics[width=0.98\linewidth]{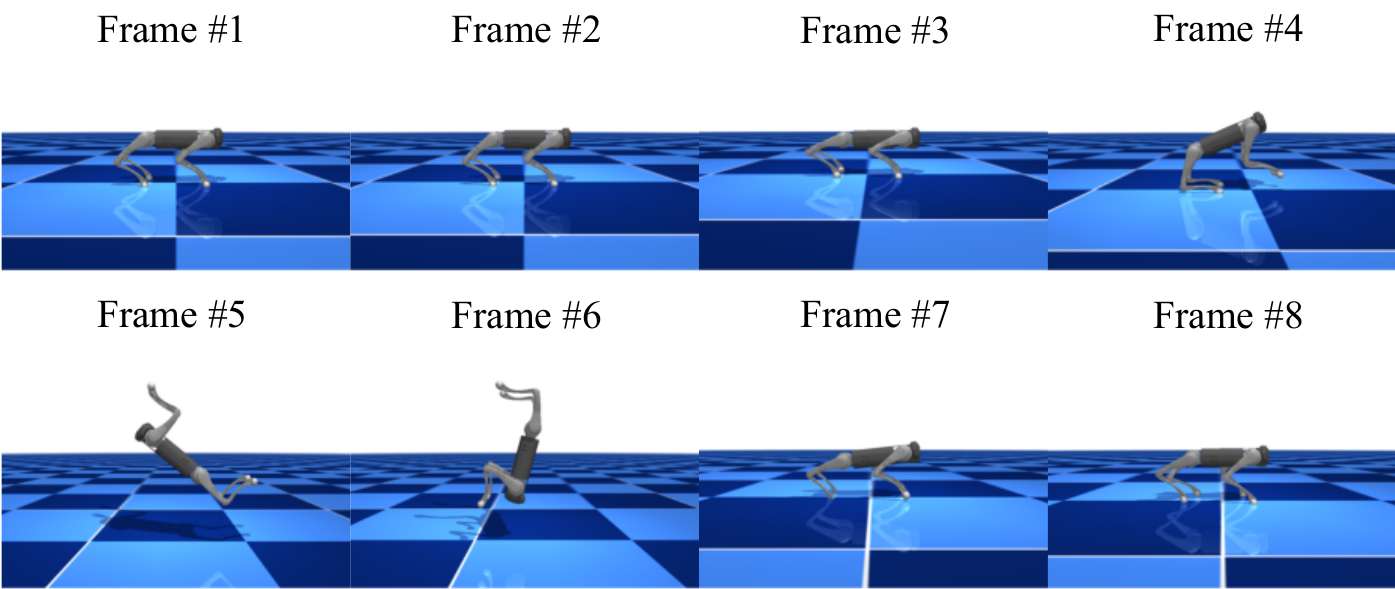}
        \label{fig:go1_task_9}
    } \\
    \vspace{0.2cm}
    \subfloat[\reviewPrev{B2 BackFlip without temporal optimization}]{
        \includegraphics[width=0.98\linewidth]{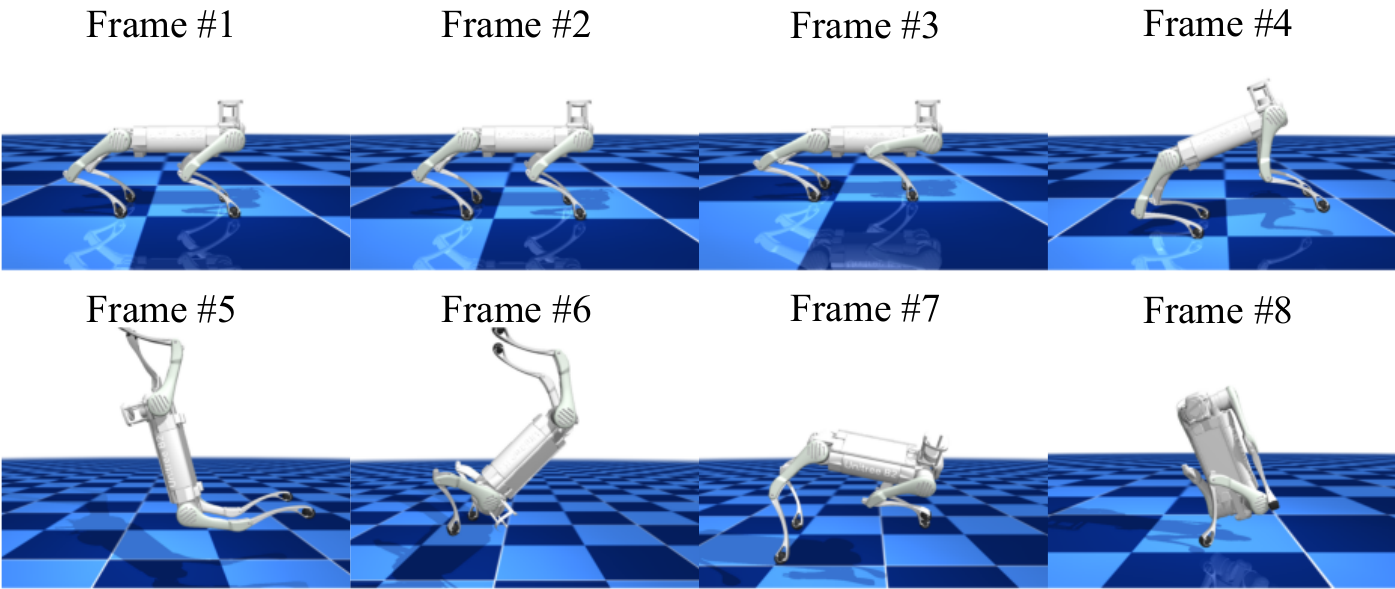}
        \label{fig:b2_task_9}
    } \\
    \vspace{0.2cm}
    \subfloat[\reviewPrev{B2 BackFlip with temporal optimization}]{
        \includegraphics[width=0.98\linewidth]{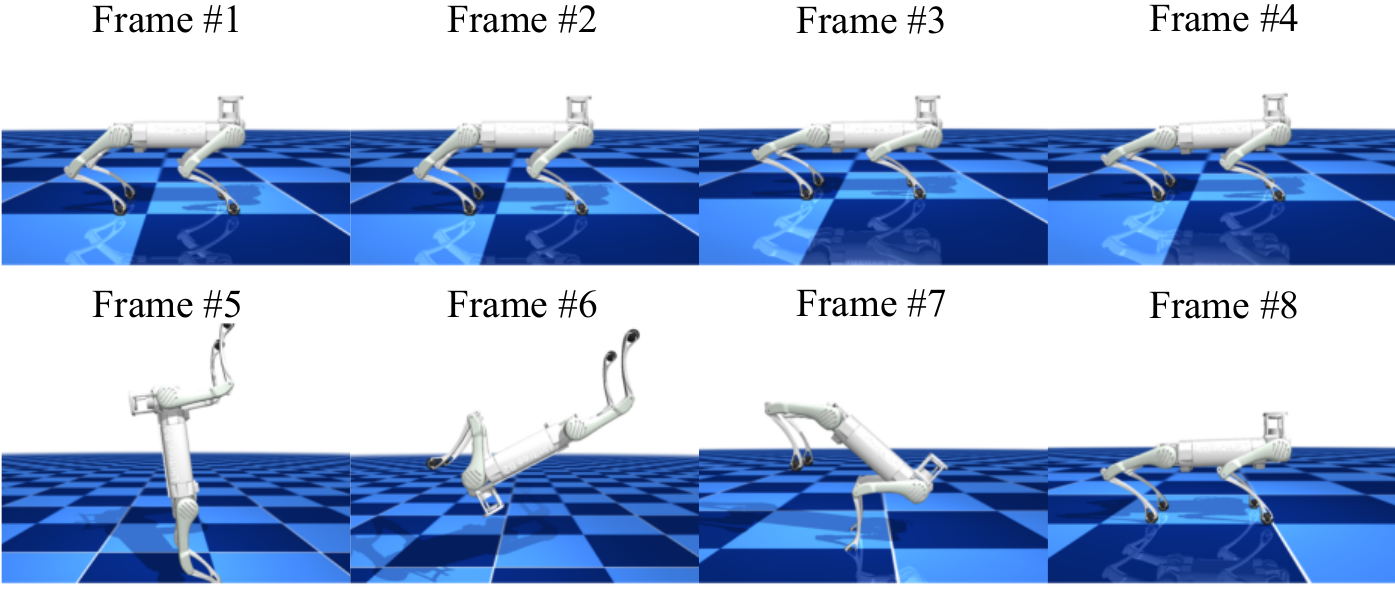}
        \label{fig:b2_task_tmr_14}
    }
    \caption{
\reviewPrev{We experiment with BackFlip to highlight the importance of temporal optimization in dynamic motions with a flight phase. (a) The original motion of Go1 is transferred to B2 (b) without temporal optimization and (c) with temporal optimization.}
    }
    \label{fig:backflip}
\end{figure}

\subsubsection{Data collection}
We collected Go1's motion data of walking forward with various gait patterns and velocities using model-based optimal control (MBOC)~\citep{tassa_synthesis_2012}. 
Specifically, the gaits include Pace (1.0 m/s), Bound (1.0 m/s), and Trot (0.5 m/s, 1.0 m/s, 1.5 m/s). 
Each motion sequence includes 3 seconds of locomotion with 0.5 seconds of standing at the beginning and end, respectively.
Furthermore, we selected forward walking motions for this experiment to evaluate the reconstruction capability by comparing the travel distance.

\subsubsection{Motion reconstruction}
We evaluate the motion reconstruction capability by removing the base trajectory $\overhorizon{\pb}{0:\maxF}$ and \reviewPrev{reconstructing it from the keypoint trajectories ${\overhorizon{\p}{0:\maxF}}$.} 
In detail, we focus on the distance traveled and measure positional differences with respect to the motions before clearing the base position.

The \Cref{table:recon} shows the distance traveled for each motion generated by SMR from whole-body motions and baseless motions. 
\reviewPrev{In detail, SideStep travel distance is measured laterally, while the rest are measured longitudinally.}
Setting the motions generated from whole-body motion as ground truth, we calculate the recovery rate.
The average recovery rate for Go1, A1, and AlienGo was 75.19\%, 74.40\%, and 78.46\%, showing that SMR can reconstruct the base trajectory regardless of configurations.

\subsubsection{Motion retargeting by reconstruction}
As illustrated in \Cref{fig:Recon}, SMR adjusts the base trajectory and subsequent local movements appropriately for the target robot. 
We evaluate how the base's travel distance changed accordingly, as shown in \Cref{table:recon}. 
On average, AlienGo exhibited a travel distance that was 13.15\% and 17.68\% longer than Go1 for the motions retargeted from whole-body motion and baseless motion, respectively. 
This increase roughly corresponds to the size difference between Go1 and AlienGo, where the horizontal lengths of Go1 and AlienGo are 540 mm and 610 mm, respectively, with a 12.96\% difference.
Similarly, A1 showed a travel distance that was 4.67\% and 6.22\% shorter than Go1. 
This decrease also roughly corresponds to the size difference, where the horizontal length of A1 is 7.41\% shorter than that of Go1.

\subsection{Evaluating flight-phase dynamic motion} \label{sec:backflip_sim}
\reviewPrev{
TMR employs temporal optimization to determine the optimal timing of motion, enabling the dynamic execution of HopTurn and SideSteps, as demonstrated in \Cref{sec:evaluate tracking}.
To further emphasize the importance of temporal optimization, we experiment with the BackFlip motion, which is even more dynamic and includes a longer flight phase. Specifically, we manually create 12 kinematic frames using the Unitree Go1 and apply temporal optimization to generate the whole-body motion, as shown in \Cref{fig:go1_task_9}.

We retarget this motion to B2 with and without temporal optimization to clarify its impact. As shown in \Cref{fig:b2_task_9}, B2 fails to perform a BackFlip without temporal optimization, resulting in a collision with the ground. 
In contrast, \Cref{fig:b2_task_tmr_14} illustrates that B2 successfully performs the BackFlip when temporal optimization is applied.
More specifically, the original motion has a duration of 
$1.48 \mathrm{s}$, whereas the temporally optimized motion for B2 extends to $2.18 \mathrm{s}$, representing a 47\% increase.
}

\begin{figure*}[tp] % <-- 핵심 변경: [tp]
    \centering
    \captionsetup[subfloat]{labelfont=normal, font=normal}
    \subfloat[{HopTurn}]{
        \includegraphics[width=0.88\textwidth]{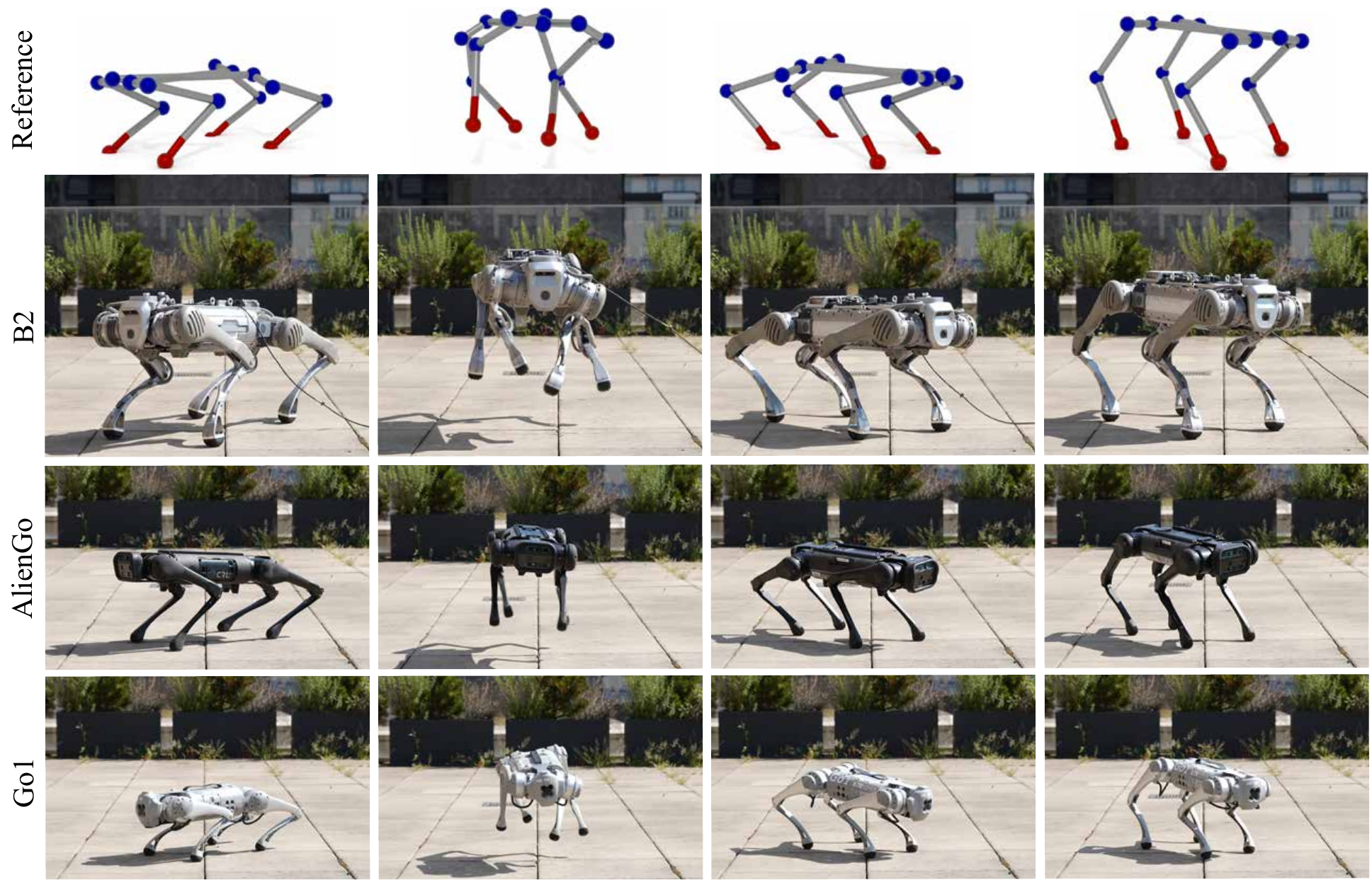}
        \label{fig:real hopturn}
    }\\[-0.5em]
    \subfloat[{SideSteps}]{
        \includegraphics[width=0.88\textwidth]{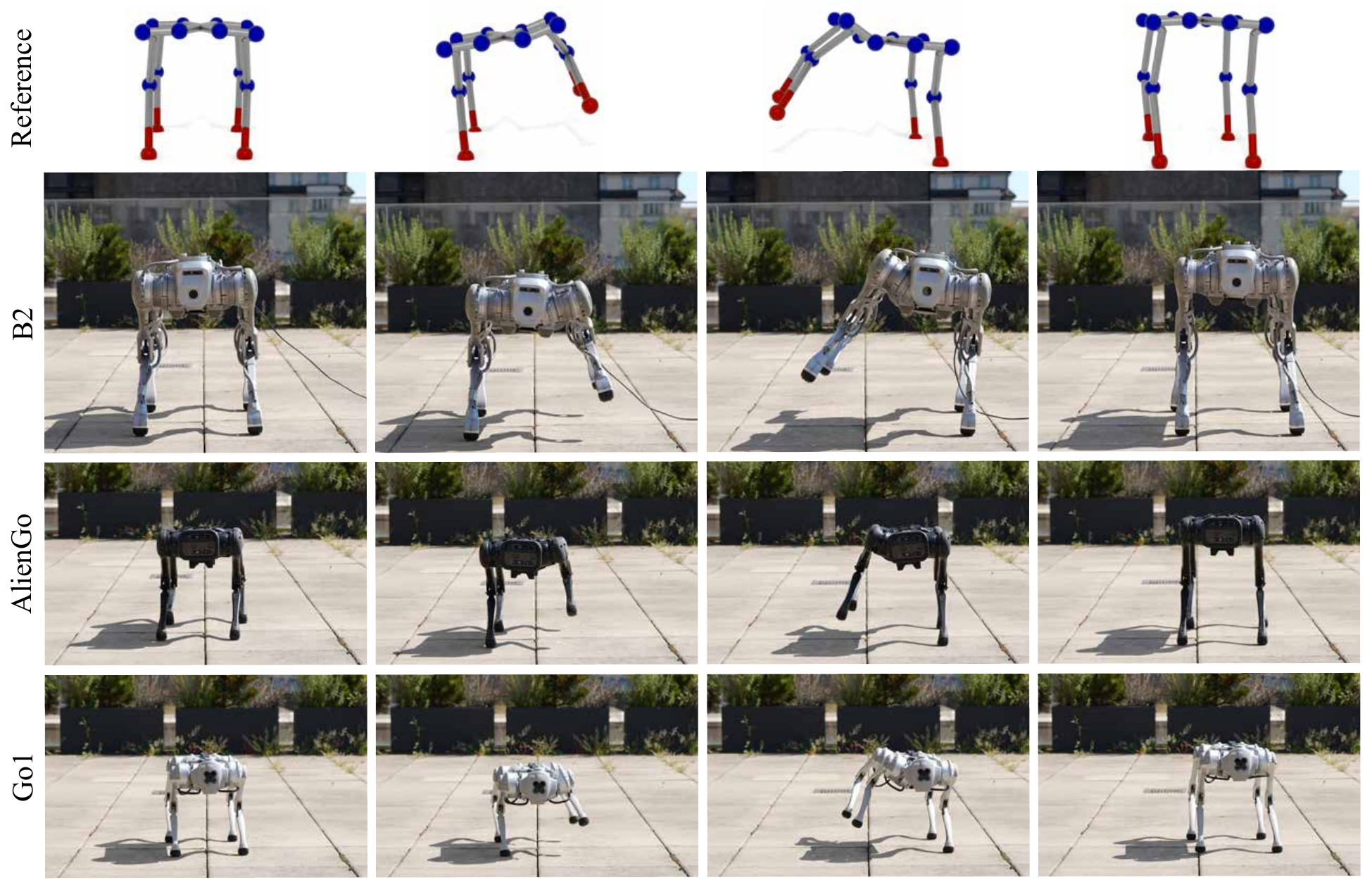}
        \label{fig:real sidesteps}
    }
    \caption{
        Real-world deployment of control policy for two motions: (a) HopTurn and (b) SideSteps. From top to bottom, the motions for the reference, Go1, AlienGo, and B2 are illustrated.
    }
    \label{fig:real experiment}
\end{figure*}

\begin{figure*}[!t]
    \centering
    \captionsetup[subfloat]{labelfont=normal, font=normal}
    \begin{minipage}{0.74\textwidth}
        \captionsetup[subfloat]{labelfont=small, font=small}
        \centering
        \includegraphics[width=0.98\linewidth]{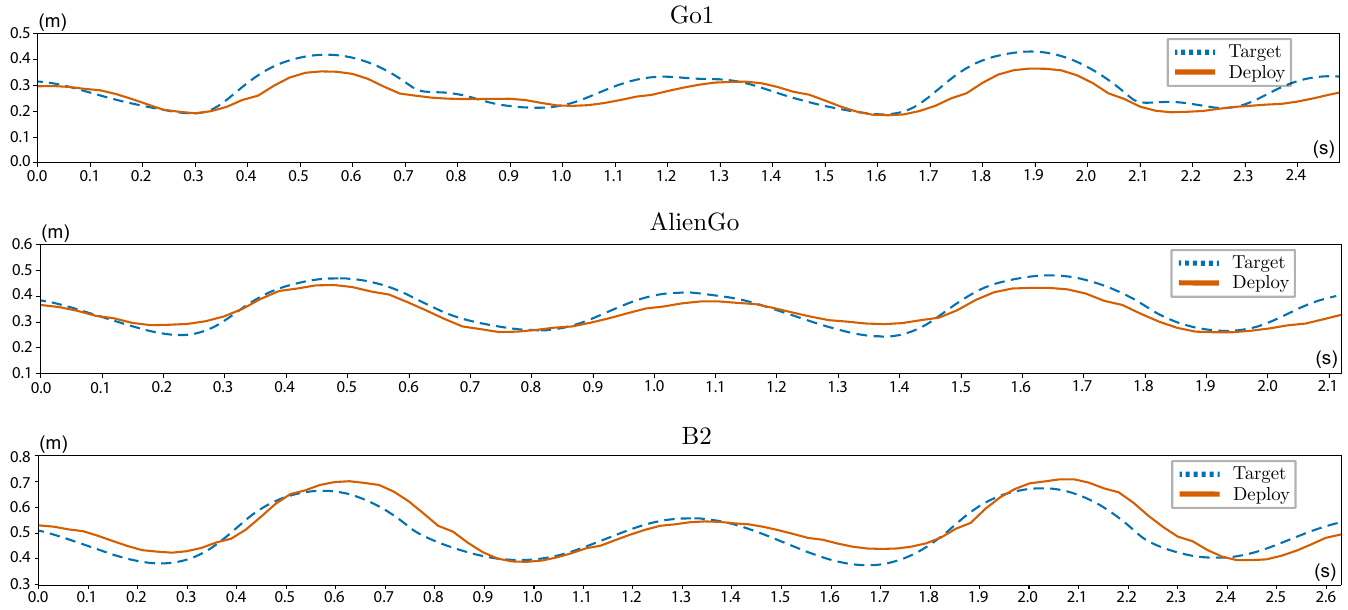}
        \subcaption{Vertical base trajectory of HopTurn}
        \label{fig:plot_z}
    \end{minipage}
    \begin{minipage}{0.21\textwidth}
        \captionsetup[subfloat]{labelfont=small, font=small}
        \centering
        \subfloat[\small{Jump height}]{
        \includegraphics[width=0.98\linewidth]{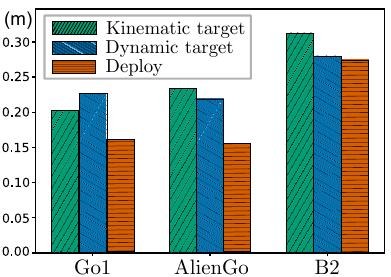}
        \label{fig:jump_height}
        }
        \vspace{0.15cm}\\
        \subfloat[\small{Jump time}]{
        \includegraphics[width=0.98\linewidth]{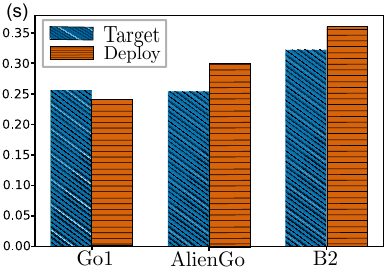}
        \label{fig:jump_time}
        } 
    \end{minipage} 
    \begin{minipage}{0.74\textwidth}
        \captionsetup[subfloat]{labelfont=small, font=small}
        \centering
        \includegraphics[width=0.98\linewidth]{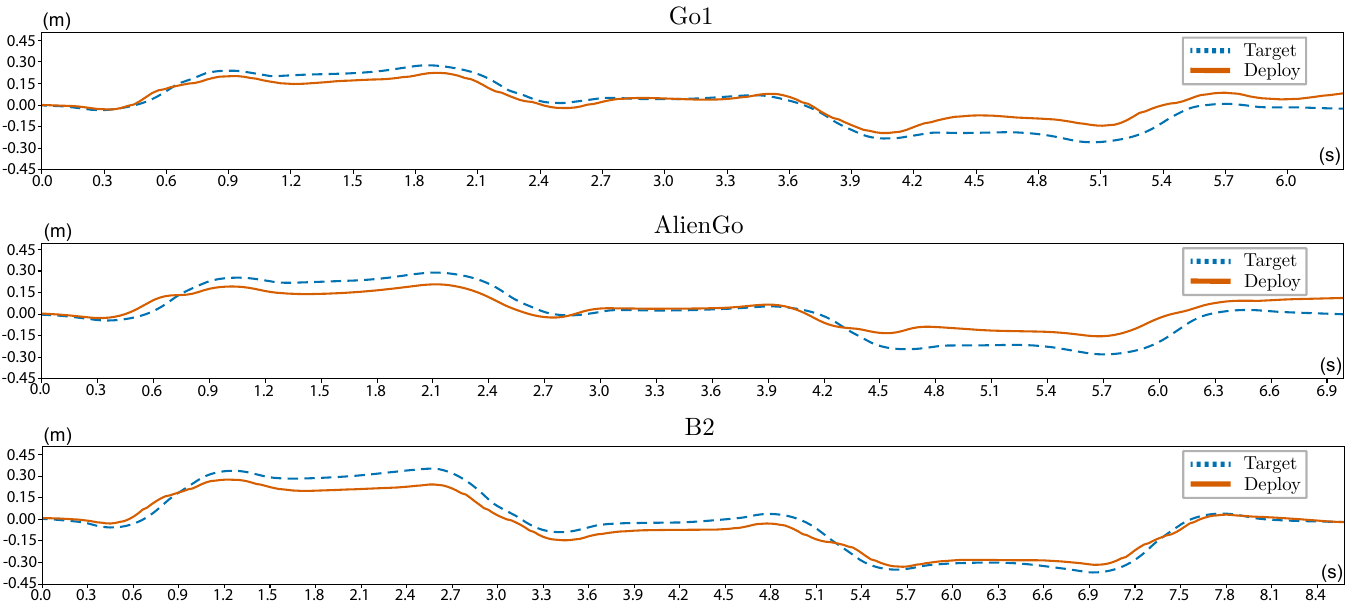}
        \subcaption{Horizontal base trajectory of SideSteps}
        \label{fig:plot_y}
    \end{minipage}
    \begin{minipage}{0.21\textwidth}
        \captionsetup[subfloat]{labelfont=small, font=small}
        \centering
        \subfloat[\small{Stepstep distance}]{
        \includegraphics[width=0.98\linewidth]{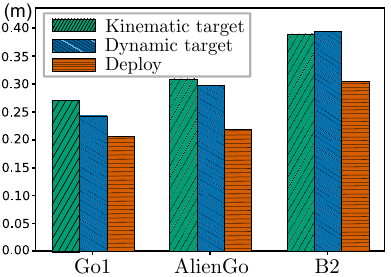}
        \label{fig:step_distance}
        }
        \vspace{0.15cm}\\
        \subfloat[\small{Stepstep time}]{
        \includegraphics[width=0.98\linewidth]{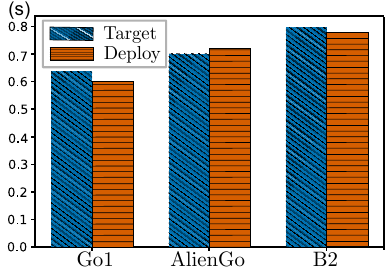}
        \label{fig:step_time}
        } 
    \end{minipage}
    \caption{
    Motion tracking results for HopTurn and SideSteps are illustrated to highlight the deformation of target motion by STMR in both spatial and temporal dimensions, as well as the tracking performance. 
    For HopTurn, (a) the vertical base trajectory, (b) the height of the first jump, and (c) the elapsed time of the first jump are illustrated. For SideSteps, (d) the horizontal base trajectory, (e) the horizontal distance of the first step, and (f) the elapsed time of the first step are shown. 
    The kinematic and dynamic targets, illustrated in (b) and (e), represent the motions refined by SMR and STMR, while Deploy indicates the motion recorded in the real world using motion capture devices.
    }
\end{figure*}

\begin{table}[!t]
    \centering
    \captionsetup{font=small}
    \footnotesize
    \setstretch{1.05}
    \caption{
    Sampling ranges for domain randomization.}
    \begin{tabular}{|c|c|c|}
        \hline
        Values & Minimum & Maximum \\
        \hline
        \hline
        Friction & 0.75 & 1.00 \\
        \hline
        Mass($\mathrm{kg}$) &  -1.0 &  1.0\\
        \hline
        Proportional Gain Multiplier & 0.9 & 1.1 \\
        \hline
        Damping Gain Multiplier & 0.9 & 1.1 \\
        \hline
        Restitution & 0.0 & 0.5 \\
        \hline
        COM displacement ($\MM$) & -100 & 100 \\
        \hline
    \end{tabular}
    \label{tab:randomization}
\end{table}

\section{Real-world Experiments}
We deployed a set of motions retargeted with our STMR method for four robot hardware platforms---Go1, Go2, AlienGo, and B2---using the learned feedback control policies.
The control policy runs at 33.33Hz, successfully responding in real-time to execute the dynamic motions of HopTurn and SideSteps on three robots, as illustrated in \Cref{fig:real experiment}, overcoming the difference in kinematic and dynamic properties\footnote{The footage from the real-world experiments can be found in the attached video.}.

To bridge the sim-to-real gap, we applied domain randomization~\citep{zhao2020sim} and introduced additional observations for the control policy.
Specifically, we randomized controller gains, mass, inertia, friction, and floor restitution, and we also applied random velocity impulses to push the robot. The details of the randomization are provided in \Cref{tab:randomization}.
In addition, we use a base state estimator~\citep{bloesch2013state} to provide the linear velocity, angular velocity, and height of the base as additional observations to the policies.
Since we have the contact schedule of the target motion in hand, we also utilize it when estimating the base state.
Introducing velocity observations helps in reducing the motion drifting over time and enables the production of more accurate jumping motions in HopTurn.

\newpage
\subsection{Analysis on motion tracking}
To examine the motion deployed in the real world, we used motion capture devices (Opti-Track Prime x22) to record the base position $\mathbf{p}_\text{b}$ and orientation $\mathbf{h}$ of the robot. 
In addition, we recorded the joint angle values $\theta$ using the motor encoders, thereby collecting the entire generalized coordinate $\mathbf{q}=[\pb, \h, \theta]$ of the robot.
Using this data, we analyzed the tracking performance and motion deformation in both spatial and temporal dimensions.

For HopTurn motion, the mean error per frame and keypoints was $30.5\mathrm{mm}$, $41.2\mathrm{mm}$, and $42.2\mathrm{mm}$ for Go1, AlienGo, and B2, respectively.
For the case with SideSteps, the mean error was $28.0\mathrm{mm}$, $35.4\mathrm{mm}$, and $34.2\mathrm{mm}$.
On average, the error across both motions was $35.3 \mathrm{mm}$, demonstrating that the closed-loop control policies accurately produce the motions on the robots, effectively overcoming uncertainties and sim-to-real gaps.

Shifting our attention to retargeted motions, we analyze how our method appropriately adjusts the motion in both the space and time domains for each robot.
\Cref{fig:plot_z} illustrates the tracking result of the base height trajectory for HopTurn with the three robots.
From the line plots, we can observe that each robot's motion execution time changed, with AlienGo being $13.95\%$ shorter than Go1 and B2 being $5.81\%$ longer than Go1.
We also measured the time and height of the first jump, which is defined as the interval between the minima and maxima of the base height trajectory. 
As shown in \Cref{fig:jump_height}, SMR retargets the source motion appropriately at the kinematic level so that the jump height increases with the robot's scale.

\begin{figure*}[!t]
    \centering
    \captionsetup[subfloat]{labelfont=normal, font=normal}
    \subfloat[\small{Estimated pose from raw video}]{
        \includegraphics[width=0.98\linewidth]{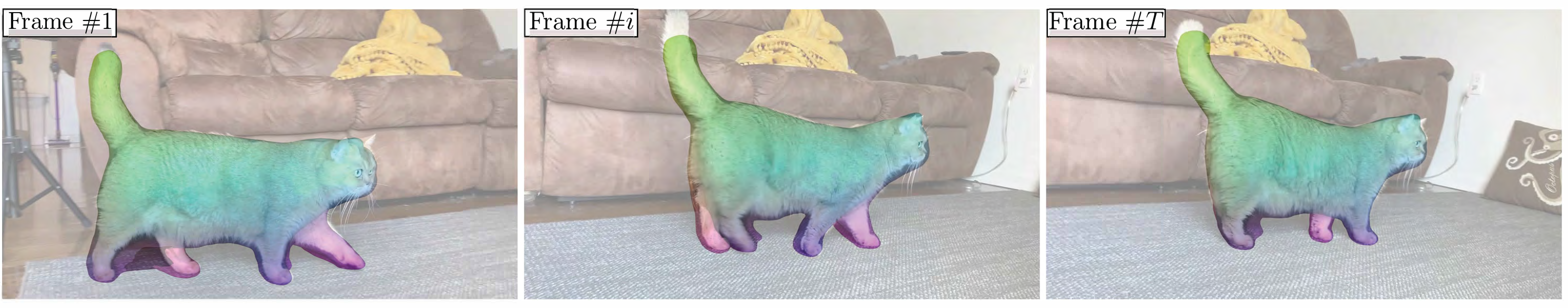}
        \label{fig:video_raw}
    } \\
    \vspace{0.2cm}
    \subfloat[\small{Reconstructed whole-body motion}]{
        \includegraphics[width=0.98\linewidth]{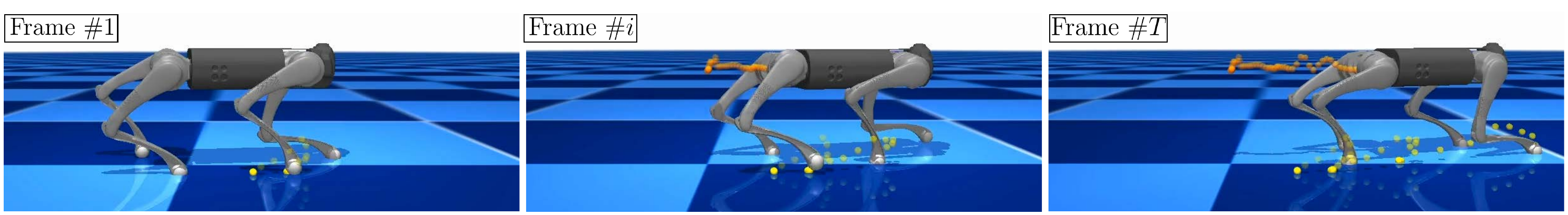}
        \label{fig:video_recon}
    } \\
    \vspace{0.2cm}
    \subfloat[\small{Real-world deployment}]{
        \includegraphics[width=0.98\linewidth]{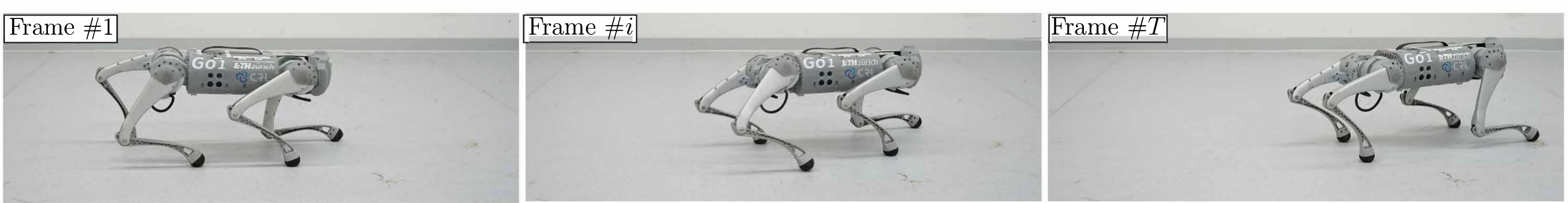}
        \label{fig:real_deploy}
    }
    \caption{The keypoint trajectories are (a) extracted using pose-estimator~\citep{banmo} and (b) reconstructed as a whole-body motion. Subsequently, the motion is utilized as a reference motion to train control policy and (c) deploy in the real world.
    }
    \label{fig:video_retarget}
\end{figure*}

% go1: 12 kg / max torque: 23.7 Nm / max speed: 30.1 rad/s / max power: 8560 W
% aliengo: 23 kg / max torque: 55 Nm / max speed: 16 rad/s / max power: 10560 W
% b2: 70 kg / max torque: 320 Nm / max speed: 23 rad/s / max power: 54720 W
%
% see https://docs.google.com/spreadsheets/d/14ln-8oBbcoJmThfLl_8IebQcLbaPJiKejcZwr5mDhvA/edit?usp=sharing
%
% reference: https://edisciplinas.usp.br/pluginfile.php/6404619/mod_resource/content/1/BoscoEJAP83jump.pdf
%
Furthermore, it is evident that STMR tailors the HopTurn motion for each robot based on its physical properties, such as weight and actuation power.
Notably, the maximum mechanical power per mass for AlienGo is $368.1\mathrm{W} / \mathrm{kg}$, which is $48.4\%$ smaller than that of Go1. %, at $713.3 \mathrm{W} \/ \mathrm{kg}$.
This results in AlienGo jumping lower than Go1 by $2.95\%$ in simulation and $3.77\%$ in the real world, despite its larger dimensions. We note that AlienGo nearly reached its maximum mechanical power during the execution of this motion.
The maximum mechanical power per mass for B2 is $9.6\%$ higher than that of Go1.
Consequently, it easily managed to jump $23.80\%$ higher than Go1.
Meanwhile, we observe the jump time increases with the robot's scale, as observed in \Cref{fig:jump_time}, with AlienGo jumping $10\%$ longer and B2 jumping $19.12\%$ longer than Go1.

For SideSteps, we focus on horizontal movement as illustrated in \Cref{fig:plot_y}.
Similar to HopTurn, each robot's total elapsed time changes where the time duration of AlienGo and B2 is extended by $10.96\%$ and $35.16\%$ compared to that of Go1.
Likewise, we compare the time and traveled distance of the first step for each robot.
As shown in \Cref{fig:step_distance}, the sidestep distance increases according to the robots' scale for all kinematic targets, dynamic targets, and real-world deployments.
To be specific, the step distance for AlienGo and B2 is larger by $14.70\%$ and $41.50\%$ than Go1.
Similarly, sidestep time also increased by $2.82\%$ and $8.45\%$ for AlienGo and B2, as illustrated in \Cref{fig:step_time}.
Again, we highlight that such deformation is caused by STMR optimizing motion in both space and time domains, resulting in successful deployment.

\subsection{Motion retargeting from videos} \label{sec:videos}
Leveraging the reconstruction capability of the proposed method, we demonstrate motion retargeting with keypoint trajectories obtained from the video pose estimator~\citep{banmo}.
Although we utilize the pose estimator that provides the base position, the estimation is considerably noisy because the global pose of the camera is unknown.
Therefore, we removed and reconstructed the base position as shown in \Cref{fig:video_recon}.
Subsequently, we temporally optimized the kinematic motion to consider the dynamic properties of the robot.
Finally, we trained a residual policy as in \Cref{sec:evaluate tracking}, where the resulting motion is illustrated in \Cref{fig:video_recon}.

Our proposed method requires contact booleans for each foot to reconstruct whole-body motion.
Therefore, we calculate foot velocity by applying finite differencing to the foot position provided by the pose estimator, followed by thresholding the foot velocity.
However, these values are noisy due to the inaccuracy of the pose estimator.
Particularly, high-frequency noise in the obtained contact leads to jerky kinematic motions.
To overcome this, we apply a low-pass filter to the foot position, making the reconstructed motion smoother.

\begin{figure}[!t]
    \centering
    \captionsetup[subfloat]{labelfont=normal, font=normal}
    \subfloat[\reviewPrev{Terrain-aware BackFlip}]{
        \includegraphics[width=0.96\linewidth]{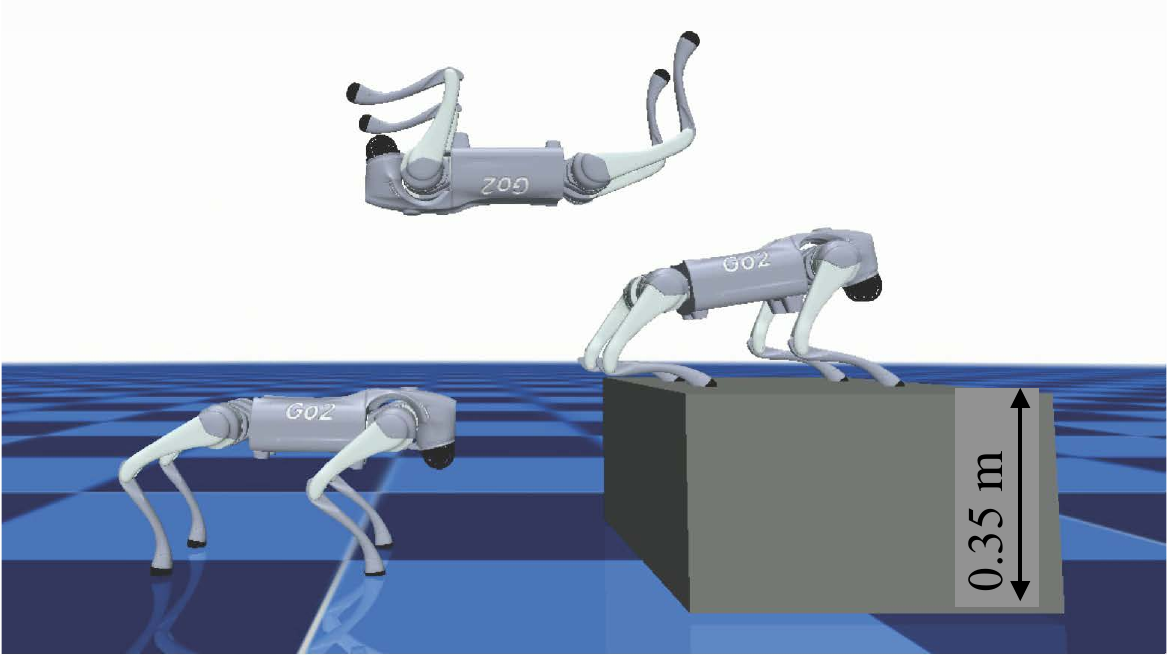}
        \label{fig:go2box3_backflip_sim}
    } \\
    \vspace{0.2cm}
    \subfloat[\reviewPrev{BackFlip deployed in the real world}]{
        \includegraphics[width=0.96\linewidth]{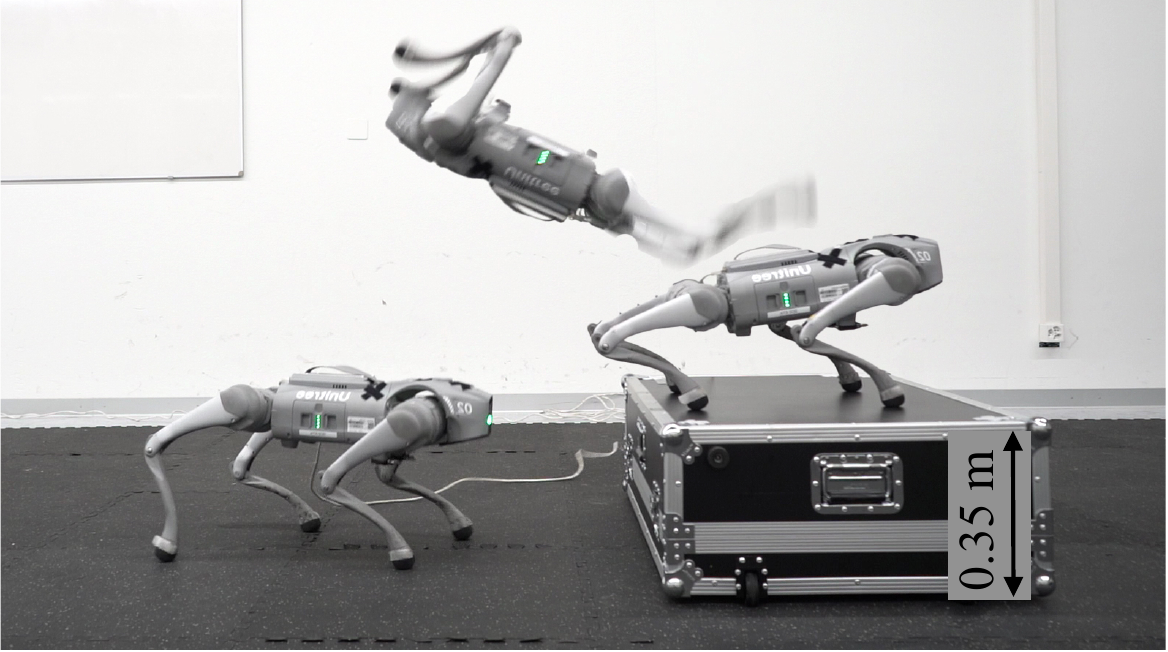}
        \label{fig:go2box3_backflip_real}
    } 
    \caption{
    \reviewPrev{(a) BackFlip is retargeted to Go2 by adapting to the terrain and ensuring the feasibility of imitation. (b) The trained policy successfully performs the BackFlip in the real world.}
    }
    \label{fig:backflip_realworld}
\end{figure}

Compared to the HopTurn and SideSteps, the motions obtained from videos involve more diverse phases, such as walking slowly followed by sharp turning.
To better handle these varying motion phases, we set the number of time segments to $\maxSeg=3$. \reviewPrev{In more detail, the tracking errors for time segments $S=1$, $S=2$, and $S=3$ were $410.8$ mm, $405.5$ mm, and $378.8$ mm, respectively, in simulation.}

\reviewPrev{\subsection{Terrain-aware retargeting of dynamic flight-phase motions}} \label{sec:backflip_realworld}
\reviewPrev{
We deploy the BackFlip motion on a real robot to demonstrate the effectiveness of our framework in refining motion both spatially and temporally. Specifically, we highlight two key capabilities: (1) spatial retargeting adapts the base trajectory to the terrain, and (2) temporal retargeting adjusts timing to ensure dynamic feasibility for real-world deployment.

Starting from the flat-ground BackFlip motion introduced in \Cref{sec:backflip_sim}, we retarget it to the Unitree Go2 robot jumping off a box with a height of $0.35\mathrm{m}$. 
To achieve this, we first generate a kinematically feasible motion for the new terrain using SMR, based on the contact schedule from the original motion.
Since SMR adapts the base trajectory to the terrain, it produces a kinematic motion that enables jumping from the box.
We then apply temporal retargeting to refine the timing and satisfy dynamic constraints.
As shown in \Cref{fig:go2box3_backflip_sim}, the resulting motion successfully performs a BackFlip in simulation.

Similar to previous experiments, we train the control policy to handle real-world uncertainties. 
A key difference, however, is that we do not provide height and linear velocity information from the state estimator, as it assumes flat terrain. 
The policy is successfully deployed in the real world and performs BackFlip, as shown in \Cref{fig:go2box3_backflip_real}.
}

\section{Limitation and Future Work}
The proposed method involves generating whole-body motion from baseless motion coherent to the contact sequence in a source motion.
This process heavily relies on accurate contact estimation, whereas wrong estimation can cause irregular movements that propagate during the construction of whole-body motions.
In particular, the estimated contact phase can be noisy as we obtain the contact boolean by thresholding the foot's velocity.
This can lead to jerky motions or failure of imitation, as robots can not follow such a quick contact transition.
In our experiments, we partially address this issue by applying a low-pass filter that regularizes high-frequency change of contact phases. 
In future work, we plan to explore more advanced contact estimation methods~\citep{zhang2024rohm} to obtain a more accurate contact sequence from the motion.

We utilized Bayesian Optimization (BO) to find the optimal temporal parameters for the TMR problem, formulated as a nested optimization problem. 
It is worth noting that optimizing over the temporal dimension is a very challenging problem, and BO offers a reasonable solution to this. 
However, BO has limitations when scaling to high-dimensional space~\citep{moriconi2020high}. Thus, it can be problematic if the source motions are long and require a larger number of time segments.
In our experiments, the duration of the motions is relatively short ($<$ 15 seconds); therefore, we set the number of time segments to a maximum of three to mitigate this issue. 
\reviewPrev{Although we can mitigate this issue by dividing motion into smaller clips, exploring more scalable optimization techniques for the TMR problem remains an area for future research.
}

%%%%%%%%%%%% Conclusion %%%%%%%%%%%%
\section{Conclusion}
This paper introduces the problem of spatio-temporal motion retargeting (STMR) that aims to generate kino-dynamically feasible motion to guide the IL process.
The STMR ensures that the transferred motions are kino-dynamically feasible for the target system. 
Furthermore, it facilitates the use of motion data with the unknown origin of reference by generating whole-body motions that closely mimic the agility and expressiveness of natural animal movements.

Our comprehensive simulation experiments demonstrate STMR's efficacy in control policy learning, which facilitates dynamic motion imitation with more accurate motion tracking performance compared to baseline methods. 
Furthermore, our method effectively preserves the contact schedule of the source motion while eliminating foot slips.
We successfully deployed the retargeted motions on four different robotic hardware platforms with varying dimensions and physical properties, highlighting the practical applicability of STMR in generating natural and dynamic motions for general quadruped robots.

Although we complemented our STMR framework with a control policy trained via reinforcement learning (RL) in this work, we note that any feedback motion controller for quadrupedal robots capable of producing whole-body movement can be used to execute the retargeted motion. 
We look forward to seeing potential extensions of this work by integrating our STMR framework with various model-based or learning-based control methods and applying it to more challenging locomotion tasks, such as bipedal locomotion or whole-body loco-manipulation tasks for legged robots.

\reviewPrev{
Moreover, STMR demonstrated the ability to retarget motions from noisy sources, such as raw video data. 
This suggests that web-scale motion datasets with physical guarantees for imitation could be realized. 
Such an approach may serve as an efficient data collection pipeline, enabling scalable and adaptable motion transfer across diverse robotic platforms. 
We leave the exploration of this direction as future work.
}

\bibliographystyle{IEEEtranN}
\bibliography{root.bib}
\end{document}